\pdfoutput=1

\documentclass[11pt]{article}

\usepackage[final]{acl}

\usepackage{times}
\usepackage{latexsym}

\usepackage[T1]{fontenc}

\usepackage[utf8]{inputenc}

\usepackage{microtype}

\usepackage{inconsolata}

\usepackage{graphicx}

\usepackage{amssymb}
\usepackage{booktabs}
\usepackage{makecell}
\usepackage{changepage}
\usepackage{svg}
\usepackage{amsmath}
\usepackage{tabularx,ragged2e}
\usepackage{tablefootnote}

\usepackage{lscape}
\usepackage{subfig}
\usepackage{graphicx}
\usepackage{float}

\usepackage{spverbatim}
\usepackage{listings}
\usepackage{comment}
\newcolumntype{L}[1]{>{\raggedright\arraybackslash}p{#1}}


\usepackage[autostyle, english = american]{csquotes}
\MakeOuterQuote{"}

\makeatletter
\newcommand{\@BIBLABEL}{\@emptybiblabel}
\newcommand{\@emptybiblabel}[1]{}
\makeatother

\usepackage[roman]{parnotes} 
\makeatletter
\def\parnoteclear{%
    \gdef\PN@text{}%
    \parnotereset }
\makeatother

\usepackage{gb4e}
\noautomath 

\makeatletter
\renewcommand{\@subex}[2]{\settowidth{\labelwidth}{#1}\itemindent\z@\labelsep#2%
         \topsep0\p@\itemsep0\p@%
         \parsep\p@\partopsep0\p@%
         \leftmargin\labelwidth%
         \ifnum\the\@xnumdepth=1
         \else\advance\leftmargin#2\relax\fi}
\makeatother

\title{GeNRe: A French Gender-Neutral Rewriting System Using Collective Nouns}

\author{Enzo Doyen \\
  LiLPa, University of Strasbourg \\
  \texttt{enzo.doyen@unistra.fr}
  \And 
  Amalia Todirascu \\
  LiLPa, University of Strasbourg\\
  \texttt{todiras@unistra.fr}}

\begin{document}
\maketitle
\begin{abstract}
A significant portion of the textual data used in the field of Natural Language Processing (NLP) exhibits gender biases, particularly due to the use of masculine generics (masculine words that are supposed to refer to mixed groups of men and women), which can perpetuate and amplify stereotypes. Gender rewriting, an NLP task that involves automatically detecting and replacing gendered forms with neutral or opposite forms (e.g., from masculine to feminine), can be employed to mitigate these biases. While such systems have been developed in a number of languages (English, Arabic, Portuguese, German, French), automatic use of gender neutralization techniques (as opposed to inclusive or gender-switching techniques) has only been studied for English. This paper presents GeNRe, the very first French gender-neutral rewriting system using collective nouns, which are gender-fixed in French. We introduce a rule-based system (RBS) tailored for the French language alongside two fine-tuned language models trained on data generated by our RBS. We also explore the use of instruct-based models to enhance the performance of our other systems and find that Claude 3 Opus combined with our dictionary achieves results close to our RBS. Through this contribution, we hope to promote the advancement of gender bias mitigation techniques in NLP for French.
\end{abstract}

\section{Introduction}

Since the 1970s, several psycholinguistic studies have shown how language influences thoughts \cite{berlinBasicColorTerms1969,kayLinguisticSignificanceMeanings1978}. Further studies examining gender in language proved that it could lead to cognitive biases \cite{jacobsonUseNonsexistPronouns1985,sczesnyCanGenderFairLanguage2016}, particularly when it comes to the use of masculine generics (MG), that is masculine words that are supposed to refer to mixed groups of men and women or to someone who gender is unknown \cite{braunCognitiveEffectsMasculine2005,gygaxGenericallyIntendedSpecifically2008,gygaxExploringOnsetMaleBiased2019,richyDemelerEffetsStereotypes2021}. For example, \citet{stahlbergNameYourFavorite2001} showed that when asked to name a celebrity in a certain field in German, respondents were more likely to give the name of a man when a MG was used in the question.

Gender bias in natural language processing (NLP) models is a critical issue that can lead to biased predictions and the amplification of bias in the training data \cite{luGenderBiasNeural2020,stanczakSurveyGenderBias2021,kotekGenderBiasStereotypes2023,ducelEvaluationAutomatiqueBiais2024}. This problem is particularly relevant for machine translation systems, which are highly susceptible to gender bias when translating between languages with different grammatical gender systems \cite{savoldiGenderBiasMachine2021,wisniewskiScreeningGenderTransfer2021,vanmassenhoveGenderBiasMachine2024}. Data augmentation, which involves balancing the amount of data for all genders in a specific language, has been suggested as a potential solution to debias NLP systems \cite{zhaoGenderBiasCoreference2018}. This led to an NLP task known as "gender rewriting," whose goal is to automatically propose alternatives to gendered sentences.

As of yet, automatic gender neutralization techniques have only been developed in English \cite{vanmassenhoveNeuTralRewriterRuleBased2021,sunTheyThemTheirs2021}. In French, the only gender rewriting system created is, to our knowledge, that of \citet{lernerINCLUREDatasetToolkit2024}, which converts MG to gender-fair forms highlighting the feminine suffix\footnote{For example, "danseurs" (male dancers or MG) and "danseuses" (female dancers) can be merged into "danseur·euses" with an interpunct (·) to refer to both men and women.}. Thus, we develop a French gender-neutral rewriting system using human collective nouns (CN), defined by \citet{lecolleNomsCollectifsHumains2019} as "nouns referring to entities comprised of groups of individuals."\footnote{In French: "nom désignant une entité composée d'un ensemble d'individus humains."} CNs have been widely discussed in the literature, especially when it comes to French \cite{flauxProposNomsCollectifs1999,lammertSemantiqueCognitionNoms2010,lammertNomsCollectifsFrancais2014,lecolleNomsCollectifsHumains2019}. Since, in French, this type of noun has a gender which does not depend upon the referent's,\footnote{For instance, "la police" ("police") refers to both policemen and policewomen.} it is an effective way of achieving gender neutralization. By focusing on gender neutralization, our work targets a writing technique that, compared to visibilization (used by \citet{lernerINCLUREDatasetToolkit2024}), tends to be less contentious among native French speakers, as it does not alter the spelling of existing words nor does it introduce non-standard or new punctuation marks to separate the feminine suffix from the base word form \cite{burnettPoliticalDimensionsGender2021}. Finally, gender neutralization challenges the binary male/female gender dichotomy and is better adapted for people whose gender falls outside of the traditional categories. Our gender-neutral rewriting system, GeNRe (\textbf{Ge}nder-\textbf{N}eutral \textbf{Re}writing System Using French Collective Nouns), is the very first gender-neutral rewriting system for French\footnote{Code and data are made publicly available on GitHub, under license CC BY-SA 4.0 \url{https://github.com/spidersouris/GeNRe}}. We present three different versions of our system: one relying on a rule-based system (RBS), one relying on fine-tuned language models and one relying on an instruct-based model.

\section{Gender in French}

In French, nouns (N) are classified as either masculine or feminine, and the gender of a noun influences the form of determiners (D), adjectives (A) and past participle verbs (V) that are syntactically related. Similarly, coreferent pronouns (P), that is pronouns referring to a previously mentioned entity, also feature the same gender. Examples \ref{ex:gender} (masculine) and \ref{ex:gender2} (feminine) highlight the syntactic differences that arise when using either an inanimate masculine noun ("courrier", \emph{mail}) and an inanimate feminine noun ("lettre", \emph{letter}).

\newcommand{\isnoun}[1]{\ensuremath{\overset{\mathbf{N}}{\text{#1}}}}
\newcommand{\isverb}[1]{\ensuremath{\overset{\mathbf{V}}{\text{#1}}}}
\newcommand{\ispron}[1]{\ensuremath{\overset{\mathbf{P}}{\text{#1}}}}
\newcommand{\isdet}[1]{\ensuremath{\overset{\mathbf{D}}{\text{#1}}}}
\newcommand{\isadj}[1]{\ensuremath{\overset{\mathbf{A}}{\text{#1}}}}

\begin{exe}
  \ex \isdet{\textbf{Le}} \isnoun{courrier} \isadj{\textbf{recommandé}} a été \isverb{\textbf{écrit}} récemment. \ispron{\textbf{Il}} est \isverb{\textbf{adressé}} à son mari.
  \newline
    \begin{small}
    (The registered [m.] mail [m.] was written [m.] recently. It [m.] is addressed [m.] to her husband.)
    \end{small}
  \label{ex:gender}\newline

  \ex \isdet{\textbf{La}} \isnoun{lettre} \isadj{\textbf{recommandée}} a été \isverb{\textbf{écrite}} récemment. \ispron{\textbf{Elle}} est \isverb{\textbf{adressée}} à son mari.
  \newline
  \begin{small}
    (The registered [f.] letter [f.] was written [f.] recently. It [f.] is addressed [f.] to her husband.)
    \end{small}
  \label{ex:gender2}
\end{exe}

The gender of human role nouns reflects the sociological gender of the referent and is motivated (for instance, "danseuse" refers to a female dancer), while gender of nouns referring to unanimated beings is arbitrary \cite{watbledLinguistiqueGenre2012}.

The masculine gender for human nouns is considered to be the "default" gender in French, and can be used in a non-specific context (in the singular form, as in Example \ref{ex:gender3}\footnote{In this example, "professeur" is considered as a masculine generics insofar as it does not refer to one specific male individual, but to any individual serving as "professor".}) or to refer to groups of people composed of both men and women (in the plural form, as in Example \ref{ex:gender4}).

\begin{exe}
    \ex \textbf{Un professeur} doit savoir faire preuve d'autorité. \newline
    \begin{small}(A professor [m.] has to know how to show authority.)\end{small}
    \label{ex:gender3}
    \ex \textbf{Les lecteurs assidus} financent le journal chaque mois. \newline
    \begin{small}(Avid readers [m.] provide financial support to the newspaper every month.)\end{small}
    \label{ex:gender4}
\end{exe}

However, the use of masculine as the default gender (masculine generics, abbreviated MG) in gender-marked languages promotes androcentric mental representations and participates in invisibilizing women \cite{jacobsonUseNonsexistPronouns1985,stahlbergNameYourFavorite2001}. A 2017 survey conducted by \citet{harrisinteractiveLecritureInclusivePopulation2017} among French speakers found that respondents tend to have male-centric representations when MG are used in the questions. Similarly, according to \citet{gabrielNeutralisingLinguisticSexism2018}, MG human nouns are more likely to be associated with male referents, and specifically highlighting the generic nature of MG does not have an effect on the biased perception of survey participants \cite{gygaxMasculineFormIts2012,rothermundRemindingMayNot2024}. Consequently, two main types of writing techniques can be used to avoid the use of MG: visibilization techniques and neutralization techniques.

Visibilization techniques seek to highlight the feminine ending of words by separating the masculine ending from the feminine one through the use of specific symbols (asterisk, interpunct: \textit{professeur·e}, as in Example \ref{ex:gender5}) or by affecting the feminine ending directly (using capital or bold letters). Neutralization techniques, on the other hand, mainly revolve around various types of words: epicene words, that is words whose form is the same for masculine and feminine, whether they may have a generic (e.g. "personne", \emph{person}, as in Example \ref{ex:gender6}) meaning or a specific (e.g. "spécialiste", \emph{specialist}) one, or words that refer to groups of people, such as CNs (e.g. "lectorat", \emph{readership}, as in Example \ref{ex:gender7}), which we use for this work. CNs have a fixed gender which is not associated with the genders of the people within that group.

\begin{exe}
    \ex \textbf{Un·e professeur·e} doit savoir faire preuve d'autorité. \newline        \begin{small}(A [m./f.] professor [m./f.] has to know how to show authority.)\end{small}
    \newline
    \label{ex:gender5}
    \ex \textbf{Une personne professeure} doit savoir faire preuve d'autorité. \newline
    \begin{small}(A person teaching has to know how to show authority.)\end{small}
    \newline
    \label{ex:gender6}
    \ex \textbf{Le lectorat assidu} finance le journal chaque mois. \newline
    \begin{small}(The avid readership provides financial support to the newspaper every month.)\end{small}
    \newline
    \label{ex:gender7}
\end{exe}


Given the impact of inclusive formulations on mitigating MG-induced gender biases \citep{koeser2015justreading,kollmayer2018breakingaway,xao2023fair}, developing a system to automatically rewrite text and reduce the prevalence of MG could be a valuable tool for data augmentation. By focusing on gender neutralization, our work aims to fill this gap and explore the potential of CNs and epicene words in promoting more inclusive language. While gender rewriting works focusing on automatic gender neutralization do exist in English \citep{heDetectPerturbNeutral2021,sunTheyThemTheirs2021,vanmassenhoveNeuTralRewriterRuleBased2021}, no such efforts have been pursued for French. Similarly, studies about gender neutralization and its application to NLP tools exist in Italian \citep{piergentiliInclusiveLanguageGenderNeutral2023} and German \citep{lardelliGenderFairPostEditingCase2023}, but gender neutralization in NLP for French has not yet been addressed.

\section{The Task of Gender Rewriting}

In order to mitigate gender bias in NLP textual data, various systems have been developed to rebalance or debias datasets by generating alternative formulations when it comes to the use of grammatical gender. While \citet{alhafniSharedTaskGender2022} were the first to define this task as "gender rewriting," similar efforts had already been pursued for Arabic \cite{habashAutomaticGenderIdentification2019}, German \cite{pomerenkeINCLUSIFYBenchmarkModel2022}, and English \cite{sunTheyThemTheirs2021}. \citet{alhafniSharedTaskGender2022} define this task as: "generating alternatives of a given Arabic sentence to match different target user gender contexts." While this definition works well for the work pursued by \citet{alhafniSharedTaskGender2022}, as they focus specifically on Arabic and create a system to switch between the masculine gender and the feminine gender, it is not universally applicable. Indeed, among the aforementioned works, several approaches to gender rewriting have been explored: \citet{habashAutomaticGenderIdentification2019} and \citet{alhafniUserCentricGenderRewriting2022} developed a system to transform Arabic sentences with masculine words into sentences with feminine equivalents, and vice versa. The system created by \citet{pomerenkeINCLUSIFYBenchmarkModel2022} provides inclusive suggestions for input sentences in German and has led to the publication of an online resource letting the user choose the type of inclusive transformation to apply. More recently, \citet{velosoRewritingApproachGender2023} also developed an inclusive gender-rewriting system for Portuguese, and \citet{lernerINCLUREDatasetToolkit2024} for French. Finally, \citet{sunTheyThemTheirs2021}, \citet{vanmassenhoveNeuTralRewriterRuleBased2021} and \citet{heDetectPerturbNeutral2021} created systems to neutralize gender in an English input sentence, but no such system exists for French. As part of this work and in order to accommodate a larger number of languages and transformation types, we reframe the initial task definition given by \citet{alhafniSharedTaskGender2022} as "generating one or more alternative sentences that either neutralize gender, adopt inclusive forms, or switch to a different gender".

\section{Methodology}

To build our automatic gender neutralization system, we propose three different approaches: a rule-based approach, a model fine-tuning approach, and an instruct-based model approach. We first create a dictionary of French CNs and their member noun counterparts, which will be used to make the necessary gender transformations. We describe this dictionary in Section \ref{sec:dict}. In Section \ref{sec:data}, we then give details about the datasets that we extracted sentences from for language model (LM) fine-tuning from RBS data, and evaluation. Finally, in Section \ref{sec:models}, we explain our experimental design with the aforementioned model types. While our work focuses specifically on French, the methodology presented below is applicable to any language which can use collective nouns as a gender-neutralizing technique (e.g., Spanish) given a dictionary of human-member nouns being already available or being created following our methodology. When it comes to syntactic changes, especially considering gender and number, those would be very similar in languages with similar inflection, such as Spanish, Italian or Portuguese. As a result, the amount of work needed to adapt our methodology to these languages specifically would be much lower compared languages encoding number or gender differently.

\subsection{Dictionary}
\label{sec:dict}

First, we manually created a dictionary with French CNs and their member noun counterparts. Three approaches were used to fill this dictionary: literature review, manual collecting and semi-automatic collecting.

\begin{description}
  \item[Literature review.] French CNs have been extensively studied in the linguistic literature. We drew on the list of 138 CNs by \citet{lecolleNomsCollectifsHumains2019}, the most exhaustive list of French CNs to our knowledge, which provided a comprehensive starting point for our dictionary. Some nouns were excluded from our dictionary due to their polysemy or restrictive semantics. For example, the CN "troupe" has multiple meanings (\emph{troop}, \emph{troupe}), and its use would require specifying the associated subdomain or group members to avoid confusion. Similarly, nouns like "duo" or "trio" have too restrictive a semantics because they only apply to groups of exactly two or three people, respectively. After careful selection, we retained 105 entries from Lecolle's list.

  \item[Manual collecting.] We empirically collected CNs from media and Internet sources over an extended period, and used Sketch Engine \citep{kilgarriff2014sketch} for corpus search. This approach allowed us to identify nouns not presented in the literature on CNs, providing a complementary perspective to the literature review. With this approach, we added 46 entries to our dictionary.

  \item[Semi-automatic collecting.] We scraped the French version of Wiktionary\footnote{\url{https://fr.wiktionary.org/wiki/}} to retrieve CNs with the suffix "-phonie", which refer to the speakers of a language (e.g. "anglophonie", \emph{English-speaking world}). We developed a Python script to generate equivalent CNs by replacing the suffix "-phonie" with "-phone" (e.g. "anglophone"). This approach enabled us to efficiently collect a set of nouns that follow a specific pattern, adding 164 entries, manually checked.
\end{description}

In total, our dictionary thus contains 315 entries. Table \ref{Table1} contains a few examples of entries in our dictionary.
 
\begin{table}[h!]
\centering
\resizebox{\columnwidth}{!}{%
\begin{tabular}{cc}
\textbf{Collective noun} & \textbf{Member noun (masc. plural)} \\
\hline
\makecell{académie \\ (\emph{academy})} & \makecell{académiciens \\ (\emph{academicians})} \\ \hline
\makecell{armée \\ (\emph{army})} & \makecell{soldats \\ (\emph{soldiers})} \\ \hline
\makecell{milice \\ (\emph{militia})} & \makecell{miliciens \\ (\emph{militiamen/women})} \\ \hline
\makecell{artillerie \\ (\emph{artillery})} & \makecell{artilleurs \\ (\emph{artillerists})} \\ \hline
\makecell{auditoire \\ (\emph{listenership})} & \makecell{auditeurs \\ (\emph{listeners})} \\ \hline
\makecell{ballet \\ (\emph{ballet})} & \makecell{danseurs \\ (\emph{dancers})} \\ \hline
\makecell{police \\ (\emph{police})} & \makecell{policiers \\ (\emph{police officers})}
\end{tabular}%
}
\caption{Collective noun-member noun dictionary overview}
\label{Table1}
\end{table}

\subsection{Datasets}
\label{sec:data}

Using our dictionary, we searched for occurrences of masculine plural member nouns in two different datasets made available for research purposes: a French Wikipedia dataset with 1.58 million texts \cite{graeloGraeloWikipedia2023}\footnote{Dataset made available here: \url{https://huggingface.co/datasets/graelo/wikipedia}. License: CC-BY-SA-3.0} and the Europarl corpus \cite{koehnEuroparlParallelCorpus2005}, which was created from the proceedings of the European Parliament and available in 21 languages, including English and French. We extracted 292,076 sentences containing member nouns from the Wikipedia dataset. The Europarl corpus was filtered to include French sentences only, and 106,878 additional sentences were extracted for model fine-tuning and evaluation (total 398,954). We publish both filtered corpora.

For the rule-based system specifically, tags were automatically added at the beginning and at the end of each member phrase in the extracted sentences, with the ID of the entry in the dictionary. This was done because member nouns may have several CN counterparts, leading to several different sentences being generated in addition to the main one. For instance, the member noun "soldats" (\emph{soldiers}) could well be replaced with CNs "armée" (\emph{army}) "bataillon" (\emph{battalion}), "infanterie" (\emph{infantry}) or "régiment" (\emph{regiment}). As we used data generated by our rule-based system for model fine-tuning (see Section \ref{sec:nm}), this was especially useful to generate all the possible variations of the input sentence, and thus increase the number of examples the models were trained on. Moreover, the use of tags also helps ensure the correct member nouns will be replaced in the input sentence, as only those that are between tags will be taken into account. Example \ref{exannot} shows how these tags are used.

\begin{exe}
    \ex Un historique permet de lister <n-126>les auteurs</n> et de consulter les modifications successives de l’article par <n-68>ses rédacteurs</n>. \newline
    \begin{small}(A history allows one to list <n-126>the authors</n> and view successive modifications to the article by <n-68>its editors</n>.)\end{small}
    \label{exannot}
\end{exe}

Finally, we created a corpus-specific evaluation dataset comprised of 250 sentences from each corpus (total 500), and we manually gender-neutralized each sentence to have gold sentences.

\subsection{Models}
\label{sec:models}

In this section, we present three different model types for gender-neutral rewriting: a rule-based model, two fine-tuned language models, and an instruct-based language model. Each model takes a different approach to the task, allowing us to compare their performance.

\subsubsection{Rule-based model}

\begin{figure}[]
	\centering 
	\includegraphics[width=0.4\textwidth]{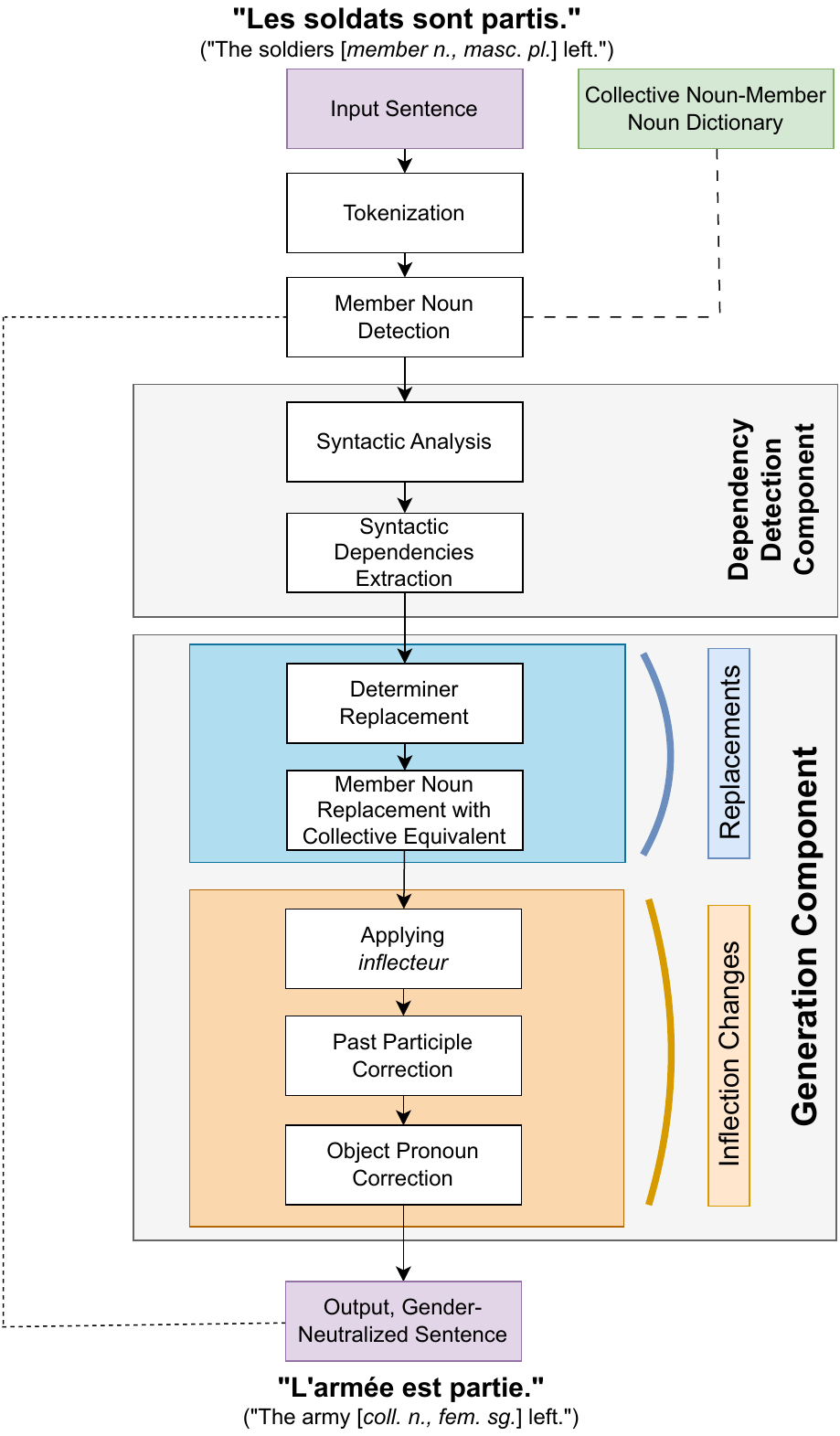}
	\caption{Rule-based model replacement pipeline overview} 
	\label{fig_mom0}%
\end{figure}

We developed an RBS to automatically apply the correct syntactic rules when converting a member noun into a CN, which leads to number and gender changes in the sentence (compare Examples \ref{ex:gender4} and \ref{ex:gender7}). The RBS consists of two main components: a syntactic dependency detection component and a generation component. Figure \ref{fig_mom0} shows an overview of the RBS pipeline.

The dependency detection component primarily relies on spaCy \cite{montaniSpaCyIndustrialstrengthNatural2024} with the \verb|fr_core_news_sm| pipeline as well as a set of rules to detect the words that are syntactically related to the member noun that needs to be replaced. We evaluated the performance of our rule-based dependency detection component in finding the correct member noun's dependencies to change compared to default spaCy-detected member noun’s dependencies (excluding punctuation), which is our baseline. We computed precision, recall and F1. Results are shown in Table \ref{table:rbsresdeps}.

The generation component replaces each member noun in the sentence with its CN counterpart found in the dictionary, adjusting the determiner, handling elision, and re-inflecting the detected dependencies using \emph{inflecteur} \cite{chuttarsingInflecteur2021}, a Python module leveraging the Delaf French morphological dictionary\footnote{\url{https://uclouvain.be/fr/instituts-recherche/ilc/cental/delaf-2-0.html}} and \emph{french-camembert-postag-model}\footnote{\url{https://huggingface.co/gilf/french-camembert-postag-model}}, a CamemBERT-based \cite{martinCamemBERTTastyFrench2020} part of speech (POS) tagging model for French. Our RBS also makes additional changes for past participles and object pronouns as these are not always being well handled by the \emph{inflecteur} Python module. If no member nouns are detected in the sentence, the original sentence will be returned instead as it is already considered gender-neutral.

We evaluate the accuracy of adjustments made by \emph{inflecteur} (as no official evaluation has been conducted) and of the additional changes made by our RBS for past participles and object pronouns. Two annotators annotated the correct inflection of syntactic dependencies found in the 500 sentences of our evaluation dataset introduced in Section \ref{sec:data} (129 dependencies for Wikipedia; 115 for Europarl: 264 in total), where $\kappa = 0.947$. \emph{inflecteur} without additional RBS changes achieves 73.01\% accuracy. With our RBS, it achieves 75.35\% accuracy (+2.34 improvement).

\subsubsection{Fine-tuned models}
\label{sec:nm}

Previous research on gender rewriting has focused on training neural models as well as fine-tuning LMs using data generated by RBS to improve task-specific performance. While some studies \cite{sunTheyThemTheirs2021,velosoRewritingApproachGender2023} showed a decrease in performance compared to RBS, \citet{vanmassenhoveNeuTralRewriterRuleBased2021} found a notable improvement of 0.27 in WER. We aim to investigate whether fine-tuning LMs can significantly improve the results of RBS, hypothesizing that the linguistic knowledge acquired by these models during training on large text corpora will help resolve errors in the training corpus and enhance results.

Two Seq2seq LMs, t5-small \cite{raffelExploringLimitsTransfer2020} and m2m100\_418M \cite{fanEnglishCentricMultilingualMachine2020}, were selected for the experiments. t5-small is a 60-million-parameter model trained on English, French, Romanian, and German. M2M100 is a multilingual model with 418 million parameters, trained on 100 languages including French. These models were chosen for their great text-to-text performance and their relatively small size, making the training process easier. Furthermore, as M2M100 had already been used by \citet{velosoRewritingApproachGender2023}, we want to compare the results we can get for our specific task. Both models were fine-tuned using our two RBS-generated corpora (Wikipedia and Europarl) containing gender-neutralized and non-gender-neutralized sentence pairs. The training dataset for each model consisted of 60,000 sentence pairs per corpus, and the validation dataset had 6,000 (10\%). Hyperparameters used for training are available in Appendix \ref{apx:hp}.

\begin{table*}[h!]
\centering
\resizebox{\textwidth}{!}{%
\begin{tabular}{c|ccc|ccc|ccc}
& \multicolumn{3}{c}{\textbf{Wikipedia}} & \multicolumn{3}{c}{\textbf{Europarl}} & \multicolumn{3}{c}{\textbf{Avg.}} \\
& \textbf{Precision} & \textbf{Recall} & \textbf{F1} & \textbf{Precision} & \textbf{Recall} & \textbf{F1} & \textbf{Precision} & \textbf{Recall} & \textbf{F1} \\
Baseline & 0.096 & 0.723 & 0.169 & 0.115 & 0.689 & 0.197 & 0.1055 & 0.706 & 0.183 \\
GeNRe-RBS & 0.773 & 0.855 & 0.812 & 0.758 & 0.813 & 0.785 & 0.7655 & 0.834 & 0.7985
\end{tabular}%
}
\caption{Results of the RBS dependency detection component per corpus and on average}
\label{table:rbsresdeps}
\end{table*}

\begin{table*}[h!]
\centering
\resizebox{\textwidth}{!}{%
\begin{tabular}{c|ccc|ccc|ccc}
& \multicolumn{3}{c}{\textbf{Wikipedia}} & \multicolumn{3}{c}{\textbf{Europarl}} & \multicolumn{3}{c}{\textbf{Avg.}} \\
& \textbf{WER (↓)} & \textbf{BLEU (↑)} & \textbf{Cosine (↑)} & \textbf{WER (↓)} & \textbf{BLEU (↑)} & \textbf{Cosine (↑)} & \textbf{WER (↓)} & \textbf{BLEU (↑)} & \textbf{Cosine (↑)} \\
Baseline & 12.611\% & 80.688 & 97.436 & 12.446\% & 82.871 & 97.008 & 12.529\% & 81.779 & 97.222 \\ 
\textbf{GeNRe-RBS} & \textbf{4.03\%} & 92.096 & \textbf{98.88} & \textbf{3.814\%} & 93.707 & 99.22 & \textbf{3.81\%} & 92.887 & \textbf{99.05} \\ 
GeNRe-T5 & 6.726\% & 87.358 & 98.508 & 4.259\% & 93.111 & 99.1 & 5.492\% & 90.234 & 98.804 \\ 
GeNRe-M2M100 & 6.566\% & 88.186 & 97.232 & 4.247\% & 93.197 & 98.992 & 5.406\% & 90.692 & 98.112 \\
Claude-BASE & 13.87\% & 80.205 & 96.532 & 10.713\% & 85.313 & 97.128 & 12.291\% & 82.759 & 96.83 \\ 
\textbf{Claude-DICT} & 4.702\% & \textbf{92.79} & 98.812 & 4.197\% & \textbf{94.247} & \textbf{99.264} & 4.45\% & \textbf{93.519} & 99.038 \\ 
Claude-CORR & 11.282\% & 84.954 & 98.092 & 8.992\% & 85.257 & 98.056 & 10.137\% & 85.25 & 98.074 \\ 
\end{tabular}%
}
\caption{Results by model type and corpus. Bold indicates the best results overall.}
\label{Table2}
\end{table*}

\subsubsection{Instruct-based model}

The rapid development of large language models (LLMs) and advances in NLP have demonstrated the ability to manipulate language models' behavior to predict text continuations and perform specific tasks without explicit training, leading to instruct-based models such as InstructGPT \cite{ouyangTrainingLanguageModels2022}, or, more recently, Llama 3 \cite{grattafioriLlama3Herd2024} or DeepSeek-V3 \cite{deepseek-aiDeepSeekV3TechnicalReport2024}. This is primarily achieved through the use of "prompts" or instructions given to the LLM \cite{liuPretrainPromptPredict2021}. Several studies \citep{nunziatini-diego-2024-implementing,bartlShowgirlsPerformersFinetuning2024,velosoRewritingApproachGender2023} have shown how instruct-based models may be used to automatically generate non-gendered-biased texts, including for translation tasks (see \citet{jourdanFairTranslateEnglishFrenchDataset2025} for English to French). However, no study has yet analyzed their capabilities for the specific task of gender rewriting in French. We chose Claude 3 Opus \verb|claude-3-opus-20240229| due to its best text generation performance at the time of the experiments \citep{anthropicClaude3Model2024} and its API being free to use during the period the experiments were conducted.\footnote{For the announcement, see \url{https://www.anthropic.com/news/claude-3-family}.}

To comprehensively evaluate the performance of Claude 3 Opus, we designed three distinct types of instructions to test its ability to generate gender-neutral texts. Corresponding prompts are available in Appendix \ref{apx:instr}.

\begin{itemize}
\item The "BASE" instruction provides a basic task description, asking the model to make the sentence inclusive by replacing MG with their CN equivalents, without explicitly specifying the replacement word.
\item The "DICT" instruction leverages our collective noun dictionary and asks the model to replace MG with their corresponding CNs, those being explicitly mentioned. There are two different versions for the "DICT" instruction: "DICT-SG", used when only one generic masculine noun with a matching CN was found in the sentence, and "DICT-PL", used when several generic masculine nouns with matching CNs were found.
\item The "CORR" instruction takes sentences generated by our RBS as input and tasks the model with correcting potential errors, such as mismatches between verb and adjective numbers and genders.
\end{itemize}

\section{Results}
\label{sec:results}

To evaluate the performance of our different rewriting models, we use the evaluation dataset presented in Section \ref{sec:data} and we leverage three evaluation metrics: Word Error Rate (WER), BLEU \cite{papineniBLEUMethodAutomatic2002} and cosine similarity. WER and BLEU have been commonly used in previous gender rewriting works. WER measures differences between the reference sentences (manually gender-rewritten) and the transformed sentences by taking into account word insertions, deletions and substitutions, and was previously used by \citet{sunTheyThemTheirs2021} and \citet{vanmassenhoveNeuTralRewriterRuleBased2021}. BLEU, on the other hand, measures the n-gram overlap between the reference and the transformed sentences ($n = 4$), and was previously used by \citet{sunTheyThemTheirs2021}, \citet{velosoRewritingApproachGender2023} and \citet{lernerINCLUREDatasetToolkit2024}. Using these metrics can help compare results from previous works (even though languages may be different) and gives an overview of the similarity between the references and the transformations (for a more precise analysis, see error labelling in Section \ref{sec:disc}). To compute WER and BLEU, we respectively use JiWER 3.0.3\footnote{\url{https://pypi.org/project/jiwer/}} and sacrebleu 2.4.2\footnote{\url{https://pypi.org/project/sacrebleu/}} \citep{post-2018-call} Python packages with default parameters. In addition, we also use cosine similarity, which has not been used in previous works. Cosine similarity evaluates the semantic closeness between sentence embeddings of the reference and transformed sentences. Recent work by \citet{piergentili2023higuyshifolks} argues that semantic similarity metrics may be ill-suited for evaluating gender-neutral rewritings as they are robust to form differences. We acknowledge this limitation and use cosine similarity not as a primary signal of neutralization quality, but as a complementary measure of meaning preservation, alongside form-sensitive metrics like BLEU and WER. For computation, we use SBERT \citep{reimers-2019-sentence-bert} and model \verb|Sentence-CamemBERT-Large|, based on CamemBERT \citep{martinCamemBERTTastyFrench2020}, for its best performances for French. The baseline corresponds to the unchanged sentence in line with previous works. Results of each model per corpus and on average are available in Table \ref{Table2}.

The RBS and Claude 3 Opus-DICT achieved the best results in our experiments, with the RBS achieving 3.81\% WER and 99.05 cosine similarity, and Claude 3 Opus-DICT achieving 93.519 BLEU. The fine-tuned models also showed mostly promising results, even though lower than the RBS and Claude 3 Opus DICT (5.492\% WER, 90.234 BLEU and 98.112 cosine similarity for T5; 5.406\% WER, 90.692 BLEU and 98.112 cosine similarity for M2M100). Comparing the two fine-tuned models, they achieved similar results, with the T5 model slightly outperforming M2M100. However, both models showed a minor decrease in performance compared to the RBS. As a result, similarly to \citet{velosoRewritingApproachGender2023} and in contrast with the findings of \citet{vanmassenhoveNeuTralRewriterRuleBased2021}, we do not find a significant improvement compared to our RBS following fine-tuning. In comparison with previous automatic gender neutralization works in English \citep{sunMitigatingGenderBias2019,vanmassenhoveNeuTralRewriterRuleBased2021}, we achieve similar (although lower for WER) difference scores between the baseline and the RBS (-11.77 WER and +9.63 BLEU for \citet{sunMitigatingGenderBias2019}; -10.947 WER for \citet{vanmassenhoveNeuTralRewriterRuleBased2021}; -8.449 WER and +11.10 BLEU for GeNRe-RBS).


\section{Discussion}
\label{sec:disc}

\definecolor{ft}{RGB}{0,97,218}
\definecolor{ins}{RGB}{204,120,92}

\begin{figure*}[h!]
    \centering
    \includegraphics[width=1\textwidth]{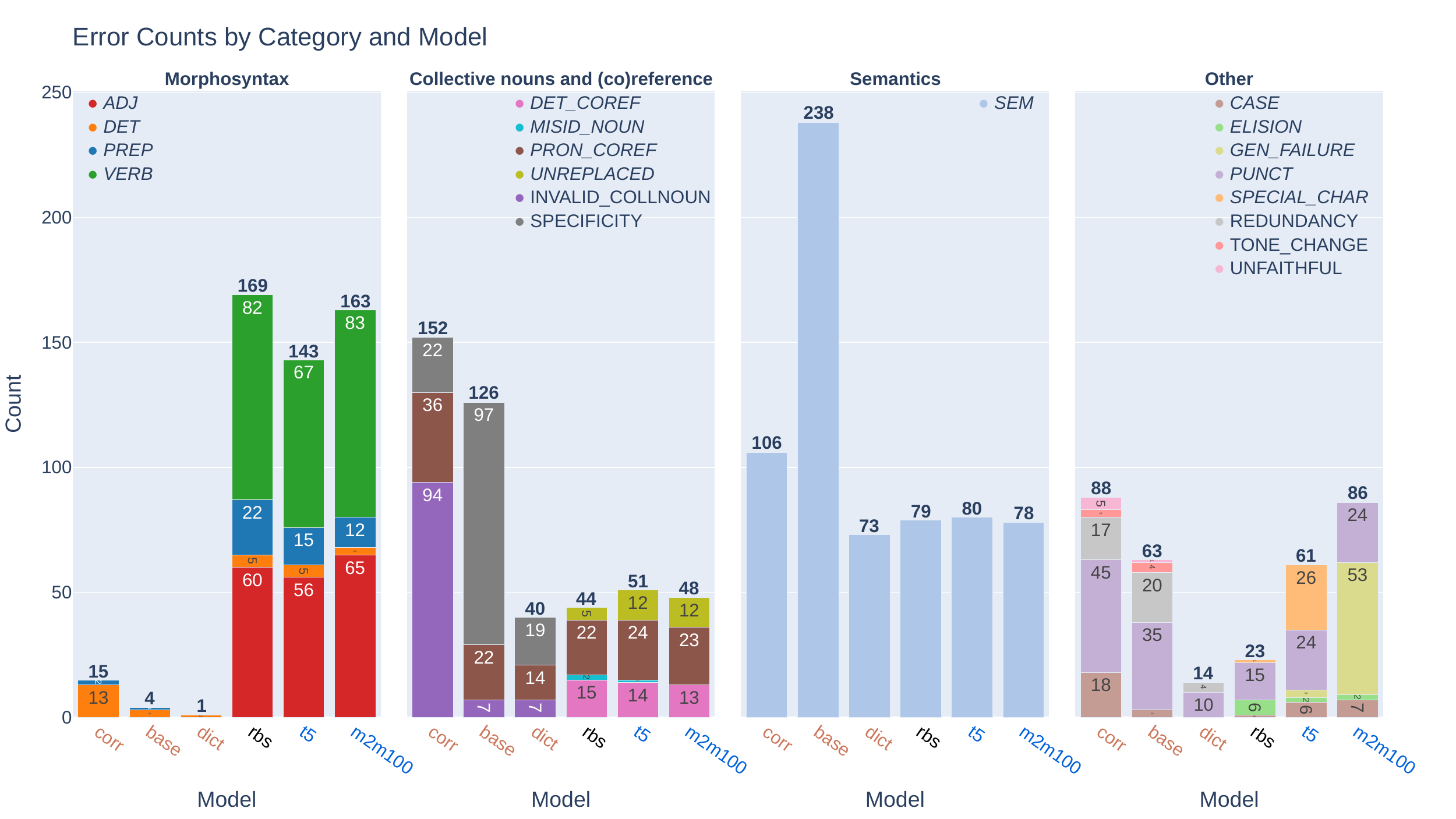}
    \caption{Error distribution for RBS, \textcolor{ft}{fine-tuned} and \textcolor{ins}{instruct-based} models}
    \label{fig_mom2}
\end{figure*}

We conduct a quantitative and qualitative analysis of errors made by each model type. This analysis is done manually for the RBS and fine-tuned models, and automatically for the instruct-based model. The latter method is chosen because LLMs tend to be significantly more permissive in their generations, resulting in a broader range of error types that would complicate manual annotation. We define four high-level categories of errors, as illustrated in Figure \ref{fig_mom2}: morphosyntax; collective nouns and (co)reference; semantics; and other.

For GeNRe-RBS, GeNRe-T5, and GeNRe-M2M100, errors were manually annotated across both corpora and verified by two annotators. Annotation details are available in Appendix \ref{apx:manualerr}. Based on the outputs from the RBS and fine-tuned models, we established 14 subcategories of errors (italicized in Figure \ref{fig_mom2}), which were then applied to each sentence generated by each model (500 sentences × 3 models). Multiple error labels can be assigned to a single sentence. Descriptions of each error type can be found in Table \ref{tab:err} in the Appendix.

For Claude 3 Opus and its instructions (BASE, DICT, CORR), we use the "LLM-as-judge" approach \cite{zheng2023judgingllmasajudgemtbenchchatbot} to automate error annotation using GPT-4o mini \cite{openaiGPT4oMiniAdvancing2024} and in-context learning \cite{brownLanguageModelsAre2020}. First, GPT-4o mini is prompted to provide a brief explanation of the error(s) in each generated sentence compared to the reference. We generate 10 different outputs and identify the most frequent explanation by comparing sentence-level embeddings using SBERT \cite{reimers-2019-sentence-bert} and Sentence-CamemBERT-Large\footnote{\url{https://huggingface.co/dangvantuan/sentence-camembert-large}}. Subcategories of errors are then assigned to each sentence based on these explanations and the manually defined taxonomy. The most frequent subcategory across the 10 generations is retained. Prompts are included in Appendix \ref{apx:autoerr}. We validate these automatically assigned subcategories by reviewing selected model outputs to ensure consistency, filtering out irrelevant or incorrectly attributed labels, and merging valid labels with those used in manual evaluation. Figure \ref{fig_mom2} shows the distribution of errors across models, grouped into the four main categories.

Morphosyntactic errors are most frequent in the RBS model (169 instances), though slightly reduced in the fine-tuned models (163 for M2M100, 143 for T5). In both cases, errors involving adjectives (ADJ) and verbs (VERB) are most prevalent. This is not surprising given that these two part-of-speech categories are the ones which require the most complex changes when transitioning from a member noun to a CN. Indeed, in French, adjectives undergo a certain number of changes when changing number or gender. Verbs can also have these same changes when used as past participles; otherwise, only number change will affect them. For instance, in Example \ref{ex3b}, the verb "veulent" (pl., \emph{want}) was correctly changed by M2M100 to "veut" (sg.) to match with the new CN "actorat" (\emph{group of actors}). Thus, despite their overall lower performance, these models show promise for correcting such errors. The instruction-tuned model variants exhibit significantly fewer such errors, reflecting their great linguistic capabilities.

For collective noun and (co)reference errors, the "INVALID\_COLLNOUN" label denotes the failure to use a valid collective noun present in the reference sentence: it is typically replaced by a MG (Example \ref{excorrinvalidcnoun}). The "SPECIFICITY" error captures shifts in noun specificity, such as changes from definite to indefinite forms.

Semantic errors are treated as a distinct category due to their frequency. All models tend to struggle with these, likely because the use of collective nouns is governed by strict semantic rules that limit their applicability in many contexts. The BASE instruction model exhibits a notably higher number of such errors (238), probably due to the increased generative freedom permitted by the prompt, which leads to the use of inappropriate or non-existent collective nouns (Example \ref{ex8}).

When it comes to other types of errors, errors observed in the fine-tuned models and different from the RBS included token generation failures (M2M100 mostly, Example \ref{exfailed}, where *"Nebski" was generated instead of "Zemski"), and incorrect generation of special characters (T5 only, as in Example \ref{exchar} where *"main-d’uvre" was generated instead of "main-d’œuvre" [\emph{labour}]). The first error might come from the multilingual aspect of the model, as it may generate words or mix tokens from other languages, while the second error is probably due to the model being mostly trained on English data. For both models, we also found cases where words were not uppercased correctly, as in Example \ref{excase}. Additional examples and their translations can be found in Appendix \ref{apx:examples}.

\begin{exe}
\ex \begin{xlist}
        \ex {[$\dots$]} le soutien apporté à la Commission à \textbf{l'actorat local} qui (\textbf{veulent\textsuperscript{RBS}} | \textbf{veut\textsuperscript{M2M100}}) participer {[$\dots$]} \newline \begin{small}({[$\dots$]} the support given to the Commmission of the \textbf{local group of actors} who (\textbf{want [pl.]\textsuperscript{RBS}} | \textbf{want [sg.]\textsuperscript{M2M100}}) to participate {[$\dots$]})\end{small}
        \label{ex3b}
        \ex Le point culminant est une attaque contre (\textbf{la rébellion cachée\textsuperscript{ORIG}} | \textbf{les rebelles cachés\textsuperscript{OPUS-CORR}}) dans les montagnes {[$\dots$]}  \newline \begin{small}(The climax is an attack against (\textbf{the rebellion hidden [fem. sg.]\textsuperscript{ORIG}} | \textbf{the rebels hidden [masc. pl.]\textsuperscript{OPUS-CORR}}) in the mountains {[$\dots$]})\end{small}
        \label{excorrinvalidcnoun}
        \ex L'armée nigérienne déplore également la perte d'un blindé type 92A, détruit par (\textbf{les soldats nigériens\textsuperscript{ORIG}} | \textbf{l'armée nigérienne\textsuperscript{RBS}}), de 6 autres véhicules armés, de deux canons de 122mm {[$\dots$]}  \newline \begin{small}(The Nigerian army also deplores the loss of a 92A armored vehicle, destroyed by (\textbf{Nigerian soldiers\textsuperscript{ORIG}} | \textbf{the Nigerian army\textsuperscript{RBS}}), as well as 6 other armed cars, two 122-mm cannons {[$\dots$]})\end{small}
        \label{exsem2}
        \ex En 1531, pour payer sa dette (\textbf{aux marchands\textsuperscript{ORIG}} | \textbf{au négoce\textsuperscript{OPUS-BASE}}) de Lübeck, le roi Gustav Vasa {[$\dots$]}  \newline \begin{small}(In 1531, to pay his debt (\textbf{to the dealers\textsuperscript{ORIG}} | \textbf{to the business\textsuperscript{OPUS-BASE}}) of Lübeck, the king Gustav Vasa {[$\dots$]})\end{small}
        \label{ex8}
        \ex Juin, Russie : le (\textbf{Zemski\textsuperscript{ORIG}} | \textbf{Nebski\textsuperscript{M2M100}}) sobor prend des décisions importantes.  \newline \begin{small}(June, Russia: the (\textbf{Zemski\textsuperscript{ORIG}} | \textbf{Nebski\textsuperscript{M2M100}}) Sobor makes important decisions.)\end{small}
        \label{exfailed}
        \ex Il est allé à Cologne, où il est devenu président de l’association de la (\textbf{main-d’œuvre\textsuperscript{ORIG}} | \textbf{main-d’uvre\textsuperscript{T5}}) {[$\dots$]}  \newline \begin{small}((He went to Cologne, where he became president of the \textbf{labour} organization {[$\dots$]})\end{small}
        \label{exchar}
        \ex (\textbf{L'\textsuperscript{ORIG}} | \textbf{l'\textsuperscript{T5}})armée arriva avec une lance à eau pour disperser les détenus.  \newline \begin{small}((\textbf{The\textsuperscript{ORIG}} | \textbf{The\textsuperscript{T5}}) army arrived with a water hose to disperse the prisoners.)\end{small}
        \label{excase}
    \end{xlist}
\end{exe}

\section{Conclusion}

Our work represents a step towards addressing gender-biased textual data in French. We make three key contributions to the task of gender rewriting in NLP: 1) a dictionary of French CNs and their corresponding member nouns, which serves as a resource for future research in this area; 2) a dataset of gender-neutralized and non-gender-neutralized sentences; and 3) a rule-based system that effectively gender-neutralizes French sentences using CNs. Our experiment combining our manually created dictionary with the Claude 3 Opus instruct-based model also shows promise for the use of such models for the task of gender rewriting. We believe that future research further exploring the capabilities of these models for that task could lead to the development of effective solutions for mitigating gender bias in other languages with collective nouns (such as Spanish) or similar gender neutralization techniques.

\section*{Limitations}

French CNs adhere to specific semantic rules, which means that their usage may not be universally applicable to all sentences, sometimes resulting in constructions that appear asemantic. This limitation is further compounded by the fact that only a small subset of these nouns is actively employed in everyday language by native speakers, which restricts their versatility and adaptability in various linguistic contexts. We however believe that they are good candidates for gender neutralization, and the development of our system may help promote a broader use of such nouns. In addition, combining our system with a contextual or semantic analysis framework could help address these issues by ensuring that the CN equivalents are both contextually relevant and semantically appropriate.

Furthermore, even though collective nouns have not been tested specifically, recent research works from \citet{spinelliNeutralNotFair2023} and \citet{tibblinMaleBiasCan2023} showed that gender neutralization appears to be less effective to counter gender biases induced by the use of MG. As previously stated, however, this writing technique is less contentious among the general population compared to others which explicitly highlight the feminine ending of words or separate it from the masculine ending.

The LLM-as-a-judge approach used for automatic instruct-based error analysis produces results that are not directly comparable with human annotation. Even if multiple steps were taken to generate consistent results, for instance by only retrieving the most common generations across multiple tries, some works have highlighted the limits of LLMs as judges \citep{gu2025surveyllmasajudge}.

Finally, this work is limited to the French language only, and the methodology we resorted to can only be used by languages with collective nouns acting as gender neutralizers (e.g., Spanish) and requires the creation of a language-specific human-member noun dictionary.

\section*{Ethics Statement}

We did not filter the datasets that were used for the development of the RBS and for fine-tuning models for harmful, hateful, inappropriate or personal content. Considering the sources used to constitute these datasets (Wikipedia and Europarl), we believe it very unlikely for those to display such type of content. Similarly, when it comes to output sentences generated by the fine-tuned models, since those were trained on replacing specific words in sentences, the generation of such content seems unlikely. As discussed in the paper, instruct-based models are more prone to reformulating input sentences: while we did not find any inappropriate content in the Claude 3 Opus-generated sentences we evaluated, LLMs may be trained on such data, which might lead to the generation of harmful or hateful content.

\section*{Acknowledgements}

We sincerely thank the reviewers for their constructive comments and suggestions, which contributed substantively in improving this paper. \\ This research was supported by the LiRiC Interdisciplinary Thematic Institute at the University of Strasbourg.

\bibliography{merged}

\begin{thebibliography}{68}
\providecommand{\natexlab}[1]{#1}

\bibitem[{Alhafni et~al.(2022{\natexlab{a}})Alhafni, Habash, and Bouamor}]{alhafniUserCentricGenderRewriting2022}
Bashar Alhafni, Nizar Habash, and Houda Bouamor. 2022{\natexlab{a}}.
\newblock \href {https://doi.org/10.18653/v1/2022.naacl-main.46} {User-{{Centric Gender Rewriting}}}.
\newblock In \emph{Proceedings of the 2022 {{Conference}} of the {{North American Chapter}} of the {{Association}} for {{Computational Linguistics}}: {{Human Language Technologies}}}, pages 618--631, Seattle, United States. Association for Computational Linguistics.

\bibitem[{Alhafni et~al.(2022{\natexlab{b}})Alhafni, Habash, Bouamor, Obeid, Alrowili, Alzeer, Alshanqiti, ElBakry, ElNokrashy, Gabr, Issam, Qaddoumi, {Vijay-Shanker}, and Zyate}]{alhafniSharedTaskGender2022}
Bashar Alhafni, Nizar Habash, Houda Bouamor, Ossama Obeid, Sultan Alrowili, Daliyah Alzeer, Khawlah~M. Alshanqiti, Ahmed ElBakry, Muhammad ElNokrashy, Mohamed Gabr, Abderrahmane Issam, Abdelrahim Qaddoumi, K.~{Vijay-Shanker}, and Mahmoud Zyate. 2022{\natexlab{b}}.
\newblock \href {https://doi.org/10.48550/arXiv.2210.12410} {The {{Shared Task}} on {{Gender Rewriting}}}.
\newblock http://arxiv.org/abs/2210.12410.
\newblock \emph{Preprint}, arXiv:2210.12410.

\bibitem[{{Anthropic}(2024)}]{anthropicClaude3Model2024}
{Anthropic}. 2024.
\newblock The {{Claude}} 3 {{Model Family}}: {{Opus}}, {{Sonnet}}, {{Haiku}}.

\bibitem[{Bartl and Leavy(2024)}]{bartlShowgirlsPerformersFinetuning2024}
Marion Bartl and Susan Leavy. 2024.
\newblock \href {https://doi.org/10.48550/arXiv.2407.04434} {From '{{Showgirls}}' to '{{Performers}}': {{Fine-tuning}} with {{Gender-inclusive Language}} for {{Bias Reduction}} in {{LLMs}}}.
\newblock \emph{Preprint}, arXiv:2407.04434.

\bibitem[{Berlin and Kay(1969)}]{berlinBasicColorTerms1969}
Brent Berlin and Paul Kay. 1969.
\newblock \emph{Basic {{Color Terms}}: {{Their Universality}} and {{Evolution}}}.
\newblock University of California Press.

\bibitem[{Braun et~al.(2005)Braun, Sczesny, and Stahlberg}]{braunCognitiveEffectsMasculine2005}
Friederike Braun, Sabine Sczesny, and Dagmar Stahlberg. 2005.
\newblock \href {https://doi.org/10.1515/comm.2005.30.1.1} {Cognitive {{Effects}} of {{Masculine Generics}} in {{German}}: {{An Overview}} of {{Empirical Findings}}}.
\newblock \emph{Communications}, 30(1):1--21.

\bibitem[{Brown et~al.(2020)Brown, Mann, Ryder, Subbiah, Kaplan, Dhariwal, Neelakantan, Shyam, Sastry, Askell, Agarwal, {Herbert-Voss}, Krueger, Henighan, Child, Ramesh, Ziegler, Wu, Winter, Hesse, Chen, Sigler, Litwin, Gray, Chess, Clark, Berner, McCandlish, Radford, Sutskever, and Amodei}]{brownLanguageModelsAre2020}
Tom~B. Brown, Benjamin Mann, Nick Ryder, Melanie Subbiah, Jared Kaplan, Prafulla Dhariwal, Arvind Neelakantan, Pranav Shyam, Girish Sastry, Amanda Askell, Sandhini Agarwal, Ariel {Herbert-Voss}, Gretchen Krueger, Tom Henighan, Rewon Child, Aditya Ramesh, Daniel~M. Ziegler, Jeffrey Wu, Clemens Winter, and 12 others. 2020.
\newblock \href {https://arxiv.org/abs/2005.14165} {Language {{Models Are Few-Shot Learners}}}.
\newblock http://arxiv.org/abs/2005.14165.
\newblock \emph{Preprint}, arXiv:2005.14165.

\bibitem[{Burnett and Pozniak(2021)}]{burnettPoliticalDimensionsGender2021}
Heather Burnett and C{\'e}line Pozniak. 2021.
\newblock \href {https://doi.org/10.1111/josl.12489} {Political dimensions of gender inclusive writing in {{Parisian}} universities}.
\newblock \emph{Journal of Sociolinguistics}, 25(5):808--831.

\bibitem[{Chuttarsing(2021)}]{chuttarsingInflecteur2021}
Adrien Chuttarsing. 2021.
\newblock Inflecteur.

\bibitem[{{DeepSeek-AI} et~al.(2024){DeepSeek-AI}, Liu, Feng, Xue, Wang, Wu, Lu, Zhao, Deng, Zhang, Ruan, Dai, Guo, Yang, Chen, Ji, Li, Lin, Dai, Luo, Hao, Chen, Li, Zhang, Bao, Xu, Wang, Zhang, Ding, Xin, Gao, Li, Qu, Cai, Liang, Guo, Ni, Li, Wang, Chen, Chen, Yuan, Qiu, Li, Song, Dong, Hu, Gao, Guan, Huang, Yu, Wang, Zhang, Xu, Xia, Zhao, Wang, Zhang, Li, Wang, Zhang, Zhang, Tang, Li, Tian, Huang, Wang, Zhang, Wang, Zhu, Chen, Du, Chen, Jin, Ge, Zhang, Pan, Wang, Xu, Zhang, Chen, Li, Lu, Zhou, Chen, Wu, Ye, Ye, Ma, Wang, Zhou, Yu, Zhou, Pan, Wang, Yun, Pei, Sun, Xiao, Zeng, Zhao, An, Liu, Liang, Gao, Yu, Zhang, Li, Jin, Wang, Bi, Liu, Wang, Shen, Chen, Zhang, Chen, Nie, Sun, Wang, Cheng, Liu, Xie, Liu, Yu, Song, Shan, Zhou, Yang, Li, Su, Lin, Li, Wang, Wei, Zhu, Zhang, Xu, Xu, Huang, Li, Zhao, Sun, Li, Wang, Yu, Zheng, Zhang, Shi, Xiong, He, Tang, Piao, Wang, Tan, Ma, Liu, Guo, Wu, Ou, Zhu, Wang, Gong, Zou, He, Zha, Xiong, Ma, Yan, Luo, You, Liu, Zhou, Wu, Ren, Ren, Sha, Fu, Xu, Huang, Zhang, Xie, Zhang,
  Hao, Gou, Ma, Yan, Shao, Xu, Wu, Zhang, Li, Gu, Zhu, Liu, Li, Xie, Song, Gao, and Pan}]{deepseek-aiDeepSeekV3TechnicalReport2024}
{DeepSeek-AI}, Aixin Liu, Bei Feng, Bing Xue, Bingxuan Wang, Bochao Wu, Chengda Lu, Chenggang Zhao, Chengqi Deng, Chenyu Zhang, Chong Ruan, Damai Dai, Daya Guo, Dejian Yang, Deli Chen, Dongjie Ji, Erhang Li, Fangyun Lin, Fucong Dai, and 181 others. 2024.
\newblock \href {https://doi.org/10.48550/arXiv.2412.19437} {{{DeepSeek-V3 Technical Report}}}.
\newblock \emph{Preprint}, arXiv:2412.19437.

\bibitem[{Ducel et~al.(2024)Ducel, N{\'e}v{\'e}ol, and Fort}]{ducelEvaluationAutomatiqueBiais2024}
Fanny Ducel, Aur{\'e}lie N{\'e}v{\'e}ol, and Kar{\"e}n Fort. 2024.
\newblock {{\'E}valuation automatique des biais de genre dans des mod{\`e}les de langue auto-r{\'e}gressifs}.
\newblock \emph{TALN 2024}.

\bibitem[{Fan et~al.(2020)Fan, Bhosale, Schwenk, Ma, {El-Kishky}, Goyal, Baines, Celebi, Wenzek, Chaudhary, Goyal, Birch, Liptchinsky, Edunov, Grave, Auli, and Joulin}]{fanEnglishCentricMultilingualMachine2020}
Angela Fan, Shruti Bhosale, Holger Schwenk, Zhiyi Ma, Ahmed {El-Kishky}, Siddharth Goyal, Mandeep Baines, Onur Celebi, Guillaume Wenzek, Vishrav Chaudhary, Naman Goyal, Tom Birch, Vitaliy Liptchinsky, Sergey Edunov, Edouard Grave, Michael Auli, and Armand Joulin. 2020.
\newblock \href {https://arxiv.org/abs/2010.11125} {Beyond {{English-Centric Multilingual Machine Translation}}}.
\newblock http://arxiv.org/abs/2010.11125.
\newblock \emph{Preprint}, arXiv:2010.11125.

\bibitem[{Flaux(1999)}]{flauxProposNomsCollectifs1999}
Nelly Flaux. 1999.
\newblock \href {https://doi.org/10.5169/SEALS-400007} {{\`A} propos des noms collectifs}.
\newblock \emph{Revue de linguistique romane}, (63):471--502.

\bibitem[{Gabriel et~al.(2018)Gabriel, Gygax, and Kuhn}]{gabrielNeutralisingLinguisticSexism2018}
Ute Gabriel, Pascal~M. Gygax, and Elisabeth~A. Kuhn. 2018.
\newblock \href {https://doi.org/10.1177/1368430218771742} {Neutralising linguistic sexism: {{Promising}} but cumbersome?}
\newblock \emph{Group Processes \& Intergroup Relations}, 21(5):844--858.

\bibitem[{{graelo}(2023)}]{graeloGraeloWikipedia2023}
{graelo}. 2023.
\newblock Graelo/wikipedia.

\bibitem[{Grattafiori et~al.(2024)Grattafiori, Dubey, Jauhri, Pandey, Kadian, {Al-Dahle}, Letman, Mathur, Schelten, Vaughan, Yang, Fan, Goyal, Hartshorn, Yang, Mitra, Sravankumar, Korenev, Hinsvark, Rao, Zhang, Rodriguez, Gregerson, Spataru, Roziere, Biron, Tang, Chern, Caucheteux, Nayak, Bi, Marra, McConnell, Keller, Touret, Wu, Wong, Ferrer, Nikolaidis, Allonsius, Song, Pintz, Livshits, Wyatt, Esiobu, Choudhary, Mahajan, {Garcia-Olano}, Perino, Hupkes, Lakomkin, AlBadawy, Lobanova, Dinan, Smith, Radenovic, Guzm{\'a}n, Zhang, Synnaeve, Lee, Anderson, Thattai, Nail, Mialon, Pang, Cucurell, Nguyen, Korevaar, Xu, Touvron, Zarov, Ibarra, Kloumann, Misra, Evtimov, Zhang, Copet, Lee, Geffert, Vranes, Park, Mahadeokar, Shah, van~der Linde, Billock, Hong, Lee, Fu, Chi, Huang, Liu, Wang, Yu, Bitton, Spisak, Park, Rocca, Johnstun, Saxe, Jia, Alwala, Prasad, Upasani, Plawiak, Li, Heafield, Stone, {El-Arini}, Iyer, Malik, Chiu, Bhalla, Lakhotia, {Rantala-Yeary}, van~der Maaten, Chen, Tan, Jenkins, Martin, Madaan, Malo,
  Blecher, Landzaat, de~Oliveira, Muzzi, Pasupuleti, Singh, Paluri, Kardas, Tsimpoukelli, Oldham, Rita, Pavlova, Kambadur, Lewis, Si, Singh, Hassan, Goyal, Torabi, Bashlykov, Bogoychev, Chatterji, Zhang, Duchenne, {\c C}elebi, Alrassy, Zhang, Li, Vasic, Weng, Bhargava, Dubal, Krishnan, Koura, Xu, He, Dong, Srinivasan, Ganapathy, Calderer, Cabral, Stojnic, Raileanu, Maheswari, Girdhar, Patel, Sauvestre, Polidoro, Sumbaly, Taylor, Silva, Hou, Wang, Hosseini, Chennabasappa, Singh, Bell, Kim, Edunov, Nie, Narang, Raparthy, Shen, Wan, Bhosale, Zhang, Vandenhende, Batra, Whitman, Sootla, Collot, Gururangan, Borodinsky, Herman, Fowler, Sheasha, Georgiou, Scialom, Speckbacher, Mihaylov, Xiao, Karn, Goswami, Gupta, Ramanathan, Kerkez, Gonguet, Do, Vogeti, Albiero, Petrovic, Chu, Xiong, Fu, Meers, Martinet, Wang, Wang, Tan, Xia, Xie, Jia, Wang, Goldschlag, Gaur, Babaei, Wen, Song, Zhang, Li, Mao, Coudert, Yan, Chen, Papakipos, Singh, Srivastava, Jain, Kelsey, Shajnfeld, Gangidi, Victoria, Goldstand, Menon, Sharma,
  Boesenberg, Baevski, Feinstein, Kallet, Sangani, Teo, Yunus, Lupu, Alvarado, Caples, Gu, Ho, Poulton, Ryan, Ramchandani, Dong, Franco, Goyal, Saraf, Chowdhury, Gabriel, Bharambe, Eisenman, Yazdan, James, Maurer, Leonhardi, Huang, Loyd, Paola, Paranjape, Liu, Wu, Ni, Hancock, Wasti, Spence, Stojkovic, Gamido, Montalvo, Parker, Burton, Mejia, Liu, Wang, Kim, Zhou, Hu, Chu, Cai, Tindal, Feichtenhofer, Gao, Civin, Beaty, Kreymer, Li, Adkins, Xu, Testuggine, David, Parikh, Liskovich, Foss, Wang, Le, Holland, Dowling, Jamil, Montgomery, Presani, Hahn, Wood, Le, Brinkman, Arcaute, Dunbar, Smothers, Sun, Kreuk, Tian, Kokkinos, Ozgenel, Caggioni, Kanayet, Seide, Florez, Schwarz, Badeer, Swee, Halpern, Herman, Sizov, Guangyi, Zhang, Lakshminarayanan, Inan, Shojanazeri, Zou, Wang, Zha, Habeeb, Rudolph, Suk, Aspegren, Goldman, Zhan, Damlaj, Molybog, Tufanov, Leontiadis, Veliche, Gat, Weissman, Geboski, Kohli, Lam, Asher, Gaya, Marcus, Tang, Chan, Zhen, Reizenstein, Teboul, Zhong, Jin, Yang, Cummings, Carvill, Shepard,
  McPhie, Torres, Ginsburg, Wang, Wu, U, Saxena, Khandelwal, Zand, Matosich, Veeraraghavan, Michelena, Li, Jagadeesh, Huang, Chawla, Huang, Chen, Garg, A, Silva, Bell, Zhang, Guo, Yu, Moshkovich, Wehrstedt, Khabsa, Avalani, Bhatt, Mankus, Hasson, Lennie, Reso, Groshev, Naumov, Lathi, Keneally, Liu, Seltzer, Valko, Restrepo, Patel, Vyatskov, Samvelyan, Clark, Macey, Wang, Hermoso, Metanat, Rastegari, Bansal, Santhanam, Parks, White, Bawa, Singhal, Egebo, Usunier, Mehta, Laptev, Dong, Cheng, Chernoguz, Hart, Salpekar, Kalinli, Kent, Parekh, Saab, Balaji, Rittner, Bontrager, Roux, Dollar, Zvyagina, Ratanchandani, Yuvraj, Liang, Alao, Rodriguez, Ayub, Murthy, Nayani, Mitra, Parthasarathy, Li, Hogan, Battey, Wang, Howes, Rinott, Mehta, Siby, Bondu, Datta, Chugh, Hunt, Dhillon, Sidorov, Pan, Mahajan, Verma, Yamamoto, Ramaswamy, Lindsay, Lindsay, Feng, Lin, Zha, Patil, Shankar, Zhang, Zhang, Wang, Agarwal, Sajuyigbe, Chintala, Max, Chen, Kehoe, Satterfield, Govindaprasad, Gupta, Deng, Cho, Virk, Subramanian,
  Choudhury, Goldman, Remez, Glaser, Best, Koehler, Robinson, Li, Zhang, Matthews, Chou, Shaked, Vontimitta, Ajayi, Montanez, Mohan, Kumar, Mangla, Ionescu, Poenaru, Mihailescu, Ivanov, Li, Wang, Jiang, Bouaziz, Constable, Tang, Wu, Wang, Wu, Gao, Kleinman, Chen, Hu, Jia, Qi, Li, Zhang, Zhang, Adi, Nam, Yu, Wang, Zhao, Hao, Qian, Li, He, Rait, DeVito, Rosnbrick, Wen, Yang, Zhao, and Ma}]{grattafioriLlama3Herd2024}
Aaron Grattafiori, Abhimanyu Dubey, Abhinav Jauhri, Abhinav Pandey, Abhishek Kadian, Ahmad {Al-Dahle}, Aiesha Letman, Akhil Mathur, Alan Schelten, Alex Vaughan, Amy Yang, Angela Fan, Anirudh Goyal, Anthony Hartshorn, Aobo Yang, Archi Mitra, Archie Sravankumar, Artem Korenev, Arthur Hinsvark, and 542 others. 2024.
\newblock \href {https://doi.org/10.48550/arXiv.2407.21783} {The {{Llama}} 3 {{Herd}} of {{Models}}}.
\newblock \emph{Preprint}, arXiv:2407.21783.

\bibitem[{Gu et~al.(2025)Gu, Jiang, Shi, Tan, Zhai, Xu, Li, Shen, Ma, Liu, Wang, Zhang, Wang, Gao, Ni, and Guo}]{gu2025surveyllmasajudge}
Jiawei Gu, Xuhui Jiang, Zhichao Shi, Hexiang Tan, Xuehao Zhai, Chengjin Xu, Wei Li, Yinghan Shen, Shengjie Ma, Honghao Liu, Saizhuo Wang, Kun Zhang, Yuanzhuo Wang, Wen Gao, Lionel Ni, and Jian Guo. 2025.
\newblock \href {https://arxiv.org/abs/2411.15594} {A survey on llm-as-a-judge}.
\newblock \emph{Preprint}, arXiv:2411.15594.

\bibitem[{Gygax et~al.(2012)Gygax, Gabriel, L{\'e}vy, Pool, Grivel, and Pedrazzini}]{gygaxMasculineFormIts2012}
Pascal Gygax, Ute Gabriel, Arik L{\'e}vy, Eva Pool, Marjorie Grivel, and Elena Pedrazzini. 2012.
\newblock \href {https://doi.org/10.1080/20445911.2011.642858} {The masculine form and its competing interpretations in {{French}}: {{When}} linking grammatically masculine role names to female referents is difficult}.
\newblock \emph{Journal of Cognitive Psychology}, 24(4):395--408.

\bibitem[{Gygax et~al.(2008)Gygax, Gabriel, Sarrasin, Oakhill, and Garnham}]{gygaxGenericallyIntendedSpecifically2008}
Pascal Gygax, Ute Gabriel, Oriane Sarrasin, Jane Oakhill, and Alan Garnham. 2008.
\newblock \href {https://doi.org/10.1080/01690960701702035} {Generically intended, but specifically interpreted: {{When}} beauticians, musicians, and mechanics are all men}.
\newblock \emph{Language and Cognitive Processes}, 23(3):464--485.

\bibitem[{Gygax et~al.(2019)Gygax, Schoenhals, L{\'e}vy, Luethold, and Gabriel}]{gygaxExploringOnsetMaleBiased2019}
Pascal~Mark Gygax, Lucie Schoenhals, Arik L{\'e}vy, Patrick Luethold, and Ute Gabriel. 2019.
\newblock \href {https://doi.org/10.3389/fpsyg.2019.01225} {Exploring the {{Onset}} of a {{Male-Biased Interpretation}} of {{Masculine Generics Among French Speaking Kindergarten Children}}}.
\newblock \emph{Frontiers in Psychology}, 10:1225.

\bibitem[{Habash et~al.(2019)Habash, Bouamor, and Chung}]{habashAutomaticGenderIdentification2019}
Nizar Habash, Houda Bouamor, and Christine Chung. 2019.
\newblock \href {https://doi.org/10.18653/v1/W19-3822} {Automatic {{Gender Identification}} and {{Reinflection}} in {{Arabic}}}.
\newblock In \emph{Proceedings of the {{First Workshop}} on {{Gender Bias}} in {{Natural Language Processing}}}, pages 155--165, Florence, Italy. Association for Computational Linguistics.

\bibitem[{{Harris Interactive}(2017)}]{harrisinteractiveLecritureInclusivePopulation2017}
{Harris Interactive}. 2017.
\newblock L'{\'e}criture inclusive : La population fran{\c c}aise conna{\^i}t-elle l'{\'e}criture inclusive ? {{Quelle}} opinion en a-t-elle ?
\newblock Technical report.

\bibitem[{He et~al.(2021)He, Majumder, and McAuley}]{heDetectPerturbNeutral2021}
Zexue He, Bodhisattwa~Prasad Majumder, and Julian McAuley. 2021.
\newblock \href {https://doi.org/10.48550/arXiv.2109.11708} {Detect and {{Perturb}}: {{Neutral Rewriting}} of {{Biased}} and {{Sensitive Text}} via {{Gradient-based Decoding}}}.
\newblock http://arxiv.org/abs/2109.11708.
\newblock \emph{Preprint}, arXiv:2109.11708.

\bibitem[{Jacobson and Insko(1985)}]{jacobsonUseNonsexistPronouns1985}
Marsha~B. Jacobson and William~R. Insko. 1985.
\newblock \href {https://doi.org/10.1007/BF00287456} {Use of {{Nonsexist Pronouns}} as a {{Function}} of {{One}}'s {{Feminist Orientation}}}.
\newblock \emph{Sex Roles}, 13(1-2):1--7.

\bibitem[{Jourdan et~al.(2025)Jourdan, Chevalier, and Favre}]{jourdanFairTranslateEnglishFrenchDataset2025}
Fanny Jourdan, Yannick Chevalier, and Cécile Favre. 2025.
\newblock \href {https://doi.org/10.48550/arXiv.2504.15941} {{{FairTranslate}}: {{An English-French Dataset}} for {{Gender Bias Evaluation}} in {{Machine Translation}} by {{Overcoming Gender Binarity}}}.
\newblock \emph{Preprint}, arXiv:2504.15941.

\bibitem[{Kay and McDaniel(1978)}]{kayLinguisticSignificanceMeanings1978}
Paul Kay and Chad~K. McDaniel. 1978.
\newblock The {{Linguistic Significance}} of the {{Meanings}} of {{Basic Color Terms}}.
\newblock \emph{Language}, 54(3):610--646.

\bibitem[{Kilgarriff et~al.(2014)Kilgarriff, Baisa, Bušta, Jakubíček, Kovář, Michelfeit, Rychlý, and Suchomel}]{kilgarriff2014sketch}
Adam Kilgarriff, Vít Baisa, Jan Bušta, Miloš Jakubíček, Vojtěch Kovář, Jan Michelfeit, Pavel Rychlý, and Vít Suchomel. 2014.
\newblock The sketch engine: ten years on.
\newblock \emph{Lexicography}, 1:7--36.

\bibitem[{Koehn(2005)}]{koehnEuroparlParallelCorpus2005}
Philipp Koehn. 2005.
\newblock Europarl: {{A Parallel Corpus}} for {{Statistical Machine Translation}}.

\bibitem[{Koeser et~al.(2015)Koeser, Kuhn, and Sczesny}]{koeser2015justreading}
Sara Koeser, Elisabeth~A. Kuhn, and Sabine Sczesny. 2015.
\newblock \href {https://doi.org/10.1177/0261927X14561119} {Just reading? how gender-fair language triggers readers' use of gender-fair forms.}
\newblock \emph{Journal of Language and Social Psychology}, 34(3):343--357.

\bibitem[{Kollmayer et~al.(2018)Kollmayer, Pfaffel, Schober, and Brandt}]{kollmayer2018breakingaway}
Marlene Kollmayer, Andreas Pfaffel, Barbara Schober, and Laura Brandt. 2018.
\newblock \href {https://doi.org/10.3389/fpsyg.2018.00985} {Breaking away from the male stereotype of a specialist: Gendered language affects performance in a thinking task}.
\newblock \emph{Frontiers in Psychology}, Volume 9 - 2018.

\bibitem[{Kotek et~al.(2023)Kotek, Dockum, and Sun}]{kotekGenderBiasStereotypes2023}
Hadas Kotek, Rikker Dockum, and David~Q. Sun. 2023.
\newblock \href {https://doi.org/10.1145/3582269.3615599} {Gender bias and stereotypes in {{Large Language Models}}}.
\newblock In \emph{Proceedings of {{The ACM Collective Intelligence Conference}}}, pages 12--24.

\bibitem[{Lammert(2010)}]{lammertSemantiqueCognitionNoms2010}
Marie Lammert. 2010.
\newblock \emph{S{\'e}mantique et Cognition : Les Noms Collectifs}.
\newblock Droz, Gen{\`e}ve.

\bibitem[{Lammert and Lecolle(2014)}]{lammertNomsCollectifsFrancais2014}
Marie Lammert and Michelle Lecolle. 2014.
\newblock Les noms collectifs en fran{\c c}ais, une vue d'ensemble.
\newblock \emph{Cahiers de lexicologie}, (105):203--222.

\bibitem[{Lardelli and Gromann(2023)}]{lardelliGenderFairPostEditingCase2023}
Manuel Lardelli and Dagmar Gromann. 2023.
\newblock \href {https://doi.org/10.5281/ZENODO.7898328} {Gender-{{Fair Post-Editing}}: {{A Case Study Beyond}} the {{Binary}}}.

\bibitem[{Lecolle(2019)}]{lecolleNomsCollectifsHumains2019}
Michelle Lecolle. 2019.
\newblock \emph{Les Noms Collectifs Humains En Fran{\c c}ais. {{Enjeux}} S{\'e}mantiques, Lexicaux et Discursifs}.
\newblock Lambert-Lucas, Universit{\'e} de Lorraine.

\bibitem[{Lerner and Grouin(2024)}]{lernerINCLUREDatasetToolkit2024}
Paul Lerner and Cyril Grouin. 2024.
\newblock {{INCLURE}}: A {{Dataset}} and {{Toolkit}} for {{Inclusive French Translation}}.

\bibitem[{Liu et~al.(2021)Liu, Yuan, Fu, Jiang, Hayashi, and Neubig}]{liuPretrainPromptPredict2021}
Pengfei Liu, Weizhe Yuan, Jinlan Fu, Zhengbao Jiang, Hiroaki Hayashi, and Graham Neubig. 2021.
\newblock \href {https://arxiv.org/abs/2107.13586} {Pre-{{Train}}, {{Prompt}}, and {{Predict}}: {{A Systematic Survey}} of {{Prompting Methods}} in {{Natural Language Processing}}}.
\newblock http://arxiv.org/abs/2107.13586.
\newblock \emph{Preprint}, arXiv:2107.13586.

\bibitem[{Lu et~al.(2020)Lu, Mardziel, Wu, Amancharla, and Datta}]{luGenderBiasNeural2020}
Kaiji Lu, Piotr Mardziel, Fangjing Wu, Preetam Amancharla, and Anupam Datta. 2020.
\newblock \href {https://doi.org/10.1007/978-3-030-62077-6_14} {Gender {{Bias}} in {{Neural Natural Language Processing}}}.
\newblock In Vivek Nigam, Tajana Ban~Kirigin, Carolyn Talcott, Joshua Guttman, Stepan Kuznetsov, Boon Thau~Loo, and Mitsuhiro Okada, editors, \emph{Logic, {{Language}}, and {{Security}}}, volume 12300, pages 189--202. Springer International Publishing, Cham.

\bibitem[{Martin et~al.(2020)Martin, Muller, Su{\'a}rez, Dupont, Romary, de~la Clergerie, Seddah, and Sagot}]{martinCamemBERTTastyFrench2020}
Louis Martin, Benjamin Muller, Pedro Javier~Ortiz Su{\'a}rez, Yoann Dupont, Laurent Romary, {\'E}ric~Villemonte de~la Clergerie, Djam{\'e} Seddah, and Beno{\^i}t Sagot. 2020.
\newblock \href {https://doi.org/10.18653/v1/2020.acl-main.645} {{{CamemBERT}}: A {{Tasty French Language Model}}}.
\newblock In \emph{Proceedings of the 58th {{Annual Meeting}} of the {{Association}} for {{Computational Linguistics}}}, pages 7203--7219.

\bibitem[{Montani et~al.(2024)Montani, Honnibal, Boyd, Landeghem, and Peters}]{montaniSpaCyIndustrialstrengthNatural2024}
Ines Montani, Matthew Honnibal, Adriane Boyd, Sofie~Van Landeghem, and Henning Peters. 2024.
\newblock \href {https://doi.org/10.5281/ZENODO.1212303} {{{spaCy}}: {{Industrial-strength Natural Language Processing}} in {{Python}}}.
\newblock Zenodo.

\bibitem[{Nunziatini and Diego(2024)}]{nunziatini-diego-2024-implementing}
Mara Nunziatini and Sara Diego. 2024.
\newblock \href {https://aclanthology.org/2024.eamt-1.48/} {Implementing gender-inclusivity in {MT} output using automatic post-editing with {LLM}s}.
\newblock In \emph{Proceedings of the 25th Annual Conference of the European Association for Machine Translation (Volume 1)}, pages 580--589, Sheffield, UK. European Association for Machine Translation (EAMT).

\bibitem[{{OpenAI}(2024)}]{openaiGPT4oMiniAdvancing2024}
{OpenAI}. 2024.
\newblock {{GPT-4o}} mini: Advancing cost-efficient intelligence.
\newblock https://openai.com/index/gpt-4o-mini-advancing-cost-efficient-intelligence/.

\bibitem[{Ouyang et~al.(2022)Ouyang, Wu, Jiang, Almeida, Wainwright, Mishkin, Zhang, Agarwal, Slama, Ray, Schulman, Hilton, Kelton, Miller, Simens, Askell, Welinder, Christiano, Leike, and Lowe}]{ouyangTrainingLanguageModels2022}
Long Ouyang, Jeff Wu, Xu~Jiang, Diogo Almeida, Carroll~L. Wainwright, Pamela Mishkin, Chong Zhang, Sandhini Agarwal, Katarina Slama, Alex Ray, John Schulman, Jacob Hilton, Fraser Kelton, Luke Miller, Maddie Simens, Amanda Askell, Peter Welinder, Paul Christiano, Jan Leike, and Ryan Lowe. 2022.
\newblock \href {https://arxiv.org/abs/2203.02155} {Training {{Language Models}} to {{Follow Instructions}} with {{Human Feedback}}}.
\newblock http://arxiv.org/abs/2203.02155.
\newblock \emph{Preprint}, arXiv:2203.02155.

\bibitem[{Papineni et~al.(2002)Papineni, Roukos, Ward, and Zhu}]{papineniBLEUMethodAutomatic2002}
Kishore Papineni, Salim Roukos, Todd Ward, and Wei-Jing Zhu. 2002.
\newblock \href {https://doi.org/10.3115/1073083.1073135} {{{BLEU}}: {{A Method}} for {{Automatic Evaluation}} of {{Machine Translation}}}.
\newblock In \emph{Proceedings of the 40th {{Annual Meeting}} on {{Association}} for {{Computational Linguistics}} - {{ACL}} '02}, Philadelphia, Pennsylvania. Association for Computational Linguistics.

\bibitem[{Piergentili et~al.(2023{\natexlab{a}})Piergentili, Fucci, Savoldi, Bentivogli, and Negri}]{piergentiliInclusiveLanguageGenderNeutral2023}
Andrea Piergentili, Dennis Fucci, Beatrice Savoldi, Luisa Bentivogli, and Matteo Negri. 2023{\natexlab{a}}.
\newblock \href {https://arxiv.org/abs/2301.10075} {From {{Inclusive Language}} to {{Gender-Neutral Machine Translation}}}.
\newblock http://arxiv.org/abs/2301.10075.
\newblock \emph{Preprint}, arXiv:2301.10075.

\bibitem[{Piergentili et~al.(2023{\natexlab{b}})Piergentili, Savoldi, Fucci, Negri, and Bentivogli}]{piergentili2023higuyshifolks}
Andrea Piergentili, Beatrice Savoldi, Dennis Fucci, Matteo Negri, and Luisa Bentivogli. 2023{\natexlab{b}}.
\newblock \href {https://arxiv.org/abs/2310.05294} {Hi guys or hi folks? benchmarking gender-neutral machine translation with the gente corpus}.
\newblock \emph{Preprint}, arXiv:2310.05294.

\bibitem[{Pomerenke(2022)}]{pomerenkeINCLUSIFYBenchmarkModel2022}
David Pomerenke. 2022.
\newblock \href {https://arxiv.org/abs/2212.02564} {{{INCLUSIFY}}: {{A Benchmark}} and a {{Model}} for {{Gender-Inclusive German}}}.
\newblock http://arxiv.org/abs/2212.02564.
\newblock \emph{Preprint}, arXiv:2212.02564.

\bibitem[{Post(2018)}]{post-2018-call}
Matt Post. 2018.
\newblock \href {https://www.aclweb.org/anthology/W18-6319} {A call for clarity in reporting {BLEU} scores}.
\newblock In \emph{Proceedings of the Third Conference on Machine Translation: Research Papers}, pages 186--191, Belgium, Brussels. Association for Computational Linguistics.

\bibitem[{Raffel et~al.(2020)Raffel, Shazeer, Roberts, Lee, Narang, Matena, Zhou, Li, and Liu}]{raffelExploringLimitsTransfer2020}
Colin Raffel, Noam Shazeer, Adam Roberts, Katherine Lee, Sharan Narang, Michael Matena, Yanqi Zhou, Wei Li, and Peter~J. Liu. 2020.
\newblock \href {https://arxiv.org/abs/1910.10683} {Exploring the {{Limits}} of {{Transfer Learning}} with a {{Unified Text-to-Text Transformer}}}.
\newblock http://arxiv.org/abs/1910.10683.
\newblock \emph{Preprint}, arXiv:1910.10683.

\bibitem[{Reimers and Gurevych(2019)}]{reimers-2019-sentence-bert}
Nils Reimers and Iryna Gurevych. 2019.
\newblock \href {http://arxiv.org/abs/1908.10084} {Sentence-bert: Sentence embeddings using siamese bert-networks}.
\newblock In \emph{Proceedings of the 2019 Conference on Empirical Methods in Natural Language Processing}. Association for Computational Linguistics.

\bibitem[{Richy and Burnett(2021)}]{richyDemelerEffetsStereotypes2021}
C{\'e}lia Richy and Heather Burnett. 2021.
\newblock \href {https://doi.org/10.4000/glad.2839} {D{\'e}m{\^e}ler les effets des st{\'e}r{\'e}otypes et le genre grammatical dans le biais masculin : Une approche exp{\'e}rimentale}.
\newblock \emph{GLAD!}, (10).

\bibitem[{Rothermund and Strack(2024)}]{rothermundRemindingMayNot2024}
Patrick Rothermund and Fritz Strack. 2024.
\newblock \href {https://doi.org/10.1177/0261927X241237739} {Reminding {{May Not Be Enough}}: {{Overcoming}} the {{Male Dominance}} of the {{Generic Masculine}}}.
\newblock \emph{Journal of Language and Social Psychology}, 43(4):468--485.

\bibitem[{Savoldi et~al.(2021)Savoldi, Gaido, Bentivogli, Negri, and Turchi}]{savoldiGenderBiasMachine2021}
Beatrice Savoldi, Marco Gaido, Luisa Bentivogli, Matteo Negri, and Marco Turchi. 2021.
\newblock \href {https://doi.org/10.1162/tacl_a_00401} {Gender {{Bias}} in {{Machine Translation}}}.
\newblock \emph{Transactions of the Association for Computational Linguistics}, 9:845--874.

\bibitem[{Sczesny et~al.(2016)Sczesny, Formanowicz, and Moser}]{sczesnyCanGenderFairLanguage2016}
Sabine Sczesny, Magda Formanowicz, and Franziska Moser. 2016.
\newblock \href {https://doi.org/10.3389/fpsyg.2016.00025} {Can {{Gender-Fair Language Reduce Gender Stereotyping}} and {{Discrimination}}?}
\newblock \emph{Frontiers in Psychology}, 7.

\bibitem[{Spinelli et~al.(2023)Spinelli, Chevrot, and Varnet}]{spinelliNeutralNotFair2023}
Elsa Spinelli, Jean-Pierre Chevrot, and L{\'e}o Varnet. 2023.
\newblock \href {https://doi.org/10.3389/fpsyg.2023.1256779} {Neutral is not fair enough: Testing the efficiency of different language gender-fair strategies}.
\newblock \emph{Frontiers in Psychology}, 14:1256779.

\bibitem[{Stahlberg et~al.(2001)Stahlberg, Sczesny, and Braun}]{stahlbergNameYourFavorite2001}
Dagmar Stahlberg, Sabine Sczesny, and Friederike Braun. 2001.
\newblock \href {https://doi.org/10.1177/0261927X01020004004} {Name {{Your Favorite Musician}}: {{Effects}} of {{Masculine Generics}} and of their {{Alternatives}} in {{German}}}.
\newblock \emph{Journal of Language and Social Psychology}, 20(4):464--469.

\bibitem[{Stanczak and Augenstein(2021)}]{stanczakSurveyGenderBias2021}
Karolina Stanczak and Isabelle Augenstein. 2021.
\newblock \href {https://arxiv.org/abs/2112.14168} {A {{Survey}} on {{Gender Bias}} in {{Natural Language Processing}}}.
\newblock http://arxiv.org/abs/2112.14168.
\newblock \emph{Preprint}, arXiv:2112.14168.

\bibitem[{Sun et~al.(2019)Sun, Gaut, Tang, Huang, ElSherief, Zhao, Mirza, Belding, Chang, and Wang}]{sunMitigatingGenderBias2019}
Tony Sun, Andrew Gaut, Shirlyn Tang, Yuxin Huang, Mai ElSherief, Jieyu Zhao, Diba Mirza, Elizabeth Belding, Kai-Wei Chang, and William~Yang Wang. 2019.
\newblock \href {https://doi.org/10.18653/v1/P19-1159} {Mitigating {{Gender Bias}} in {{Natural Language Processing}}: {{Literature Review}}}.
\newblock \emph{Proceedings of the 57th Annual Meeting of the Association for Computational Linguistics}, page~10.

\bibitem[{Sun et~al.(2021)Sun, Webster, Shah, Wang, and Johnson}]{sunTheyThemTheirs2021}
Tony Sun, Kellie Webster, Apu Shah, William~Yang Wang, and Melvin Johnson. 2021.
\newblock \href {https://arxiv.org/abs/2102.06788} {They, {{Them}}, {{Theirs}}: {{Rewriting}} with {{Gender-Neutral English}}}.
\newblock \emph{Preprint}, arXiv:2102.06788.

\bibitem[{Tibblin et~al.(2023)Tibblin, Granfeldt, Van De~Weijer, and Gygax}]{tibblinMaleBiasCan2023}
Julia Tibblin, Jonas Granfeldt, Joost Van De~Weijer, and Pascal Gygax. 2023.
\newblock \href {https://doi.org/10.5070/G60111267} {The male bias can be attenuated in reading: On the resolution of anaphoric expressions following gender-fair forms in {{French}}}.
\newblock \emph{Glossa Psycholinguistics}, 2(1).

\bibitem[{Vanmassenhove(2024)}]{vanmassenhoveGenderBiasMachine2024}
Eva Vanmassenhove. 2024.
\newblock \href {https://doi.org/10.48550/arXiv.2401.10016} {Gender {{Bias}} in {{Machine Translation}} and {{The Era}} of {{Large Language Models}}}.
\newblock \emph{Preprint}, arXiv:2401.10016.

\bibitem[{Vanmassenhove et~al.(2021)Vanmassenhove, Emmery, and Shterionov}]{vanmassenhoveNeuTralRewriterRuleBased2021}
Eva Vanmassenhove, Chris Emmery, and Dimitar Shterionov. 2021.
\newblock \href {https://doi.org/10.48550/arXiv.2109.06105} {{{NeuTral Rewriter}}: {{A Rule-Based}} and {{Neural Approach}} to {{Automatic Rewriting}} into {{Gender-Neutral Alternatives}}}.
\newblock http://arxiv.org/abs/2109.06105.
\newblock \emph{Preprint}, arXiv:2109.06105.

\bibitem[{Veloso et~al.(2023)Veloso, Coheur, and Ribeiro}]{velosoRewritingApproachGender2023}
Leonor Veloso, Luisa Coheur, and Rui Ribeiro. 2023.
\newblock \href {https://doi.org/10.18653/v1/2023.findings-emnlp.585} {A {{Rewriting Approach}} for {{Gender Inclusivity}} in {{Portuguese}}}.
\newblock In \emph{Findings of the {{Association}} for {{Computational Linguistics}}: {{EMNLP}} 2023}, pages 8747--8759, Singapore. Association for Computational Linguistics.

\bibitem[{Watbled(2012)}]{watbledLinguistiqueGenre2012}
Jean-Philippe Watbled. 2012.
\newblock Linguistique du genre.
\newblock \emph{L'Harmattan}, pages 167--179.

\bibitem[{Wisniewski et~al.(2021)Wisniewski, Zhu, Ballier, and Yvon}]{wisniewskiScreeningGenderTransfer2021}
Guillaume Wisniewski, Lichao Zhu, Nicolas Ballier, and Fran{\c c}ois Yvon. 2021.
\newblock \href {https://doi.org/10.18653/v1/2021.blackboxnlp-1.24} {Screening {{Gender Transfer}} in {{Neural Machine Translation}}}.
\newblock In \emph{Proceedings of the {{Fourth BlackboxNLP Workshop}} on {{Analyzing}} and {{Interpreting Neural Networks}} for {{NLP}}}, pages 311--321.

\bibitem[{Xiao et~al.(2023)Xiao, Strickland, and Peperkamp}]{xao2023fair}
Hualin Xiao, Brent Strickland, and Sharon Peperkamp. 2023.
\newblock \href {https://doi.org/10.1177/0261927X221084643} {How fair is gender-fair language? insights from gender ratio estimations in french}.
\newblock \emph{Journal of Language and Social Psychology}, 42(1):82--106.

\bibitem[{Zhao et~al.(2018)Zhao, Wang, Yatskar, Ordonez, and Chang}]{zhaoGenderBiasCoreference2018}
Jieyu Zhao, Tianlu Wang, Mark Yatskar, Vicente Ordonez, and Kai-Wei Chang. 2018.
\newblock \href {https://doi.org/10.48550/arXiv.1804.06876} {Gender {{Bias}} in {{Coreference Resolution}}: {{Evaluation}} and {{Debiasing Methods}}}.
\newblock \emph{Preprint}, arXiv:1804.06876.

\bibitem[{Zheng et~al.(2023)Zheng, Chiang, Sheng, Zhuang, Wu, Zhuang, Lin, Li, Li, Xing, Zhang, Gonzalez, and Stoica}]{zheng2023judgingllmasajudgemtbenchchatbot}
Lianmin Zheng, Wei-Lin Chiang, Ying Sheng, Siyuan Zhuang, Zhanghao Wu, Yonghao Zhuang, Zi~Lin, Zhuohan Li, Dacheng Li, Eric~P. Xing, Hao Zhang, Joseph~E. Gonzalez, and Ion Stoica. 2023.
\newblock \href {https://arxiv.org/abs/2306.05685} {Judging llm-as-a-judge with mt-bench and chatbot arena}.
\newblock \emph{Preprint}, arXiv:2306.05685.

\end{thebibliography}

\clearpage

\appendix
\renewcommand\thexnumi{F\arabic{xnumi}}

\section{Fine-Tuning Details}
\label{apx:hp}

Models were trained on a single NVIDIA RTX 4090 GPU. Training time took approximately 3 hours for each model.

\subsection{GeNRe-T5}

\begin{verbatim}
BATCH_SIZE = 48
NUM_PROCS = 16
EPOCHS = 5
LEARNING_RATE = 0.0005
WEIGHT_DECAY = 0.02
\end{verbatim}

\subsection{GeNRe-M2M100}

\begin{verbatim}
BATCH_SIZE = 8
NUM_PROCS = 16
EPOCHS = 5
LEARNING_RATE = 0.0005
WEIGHT_DECAY = 0.02
\end{verbatim}

\section{Instruction Details}
\label{apx:instr}

\subsection{Instruct-Based Model Hyperparameters}

\begin{verbatim}
model="claude-3-opus-20240229",
temperature=0,
messages=[
    {"role": "user",
    "content": f"{message}"},
    {"role": "assistant",
    "content": "Here is the
    output sentence:"}
]
\end{verbatim}

\subsection{Types of Instructions}

Table \ref{table:instruc} contains the different types of instructions given to Claude 3 Opus as well as their respective content.

"EXAMPLES" refers to the few-shot sentences given to the instruct-based model. See Tables \ref{table:instruc-anx-bd} and \ref{table:instruc-anx-c} for more information.

"ORIGINAL SENTENCE" is replaced with the sentence containing one or several masculine generic nouns that we want to replace with their collective counterparts. It is part of the prompt in a similar way to the example sentences so that the instruct-based model is guided towards generating the final, gender-neutralized sentence.

\begin{table}[h!]
\centering
\resizebox{\columnwidth}{!}{%
\begin{tabular}{@{}cc@{}}
\hline
\textbf{Instruction Type} & \textbf{Content}                                                                                                                                                                                                                                                                                                  \\ \hline
\makecell{BASE}                        & \makecell{Make this French sentence inclusive \\ by replacing generic masculine nouns \\ with their French collective noun equivalents. \\ Generate the final sentence only \\ without any comments nor notes. \\ \{EXAMPLES\} \\ \{ORIGINAL SENTENCE\} →}                                            \\ \hline
\makecell{DICT-SG}                     & \makecell{Make this French sentence inclusive \\ by replacing generic masculine noun \{NM\} \\ with its respective French collective noun equivalent \{NCOLL\}. \\ Generate the final sentence only \\ without any comments nor notes. \\ \{EXAMPLES\} \\ \{ORIGINAL SENTENCE\} →}
\\ \hline
\makecell{DICT-PL}                     & \makecell{Make this French sentence inclusive \\ by replacing generic masculine nouns \{NM1, NM2, …\} \\ with their respective French collective noun equivalents \{NCOLL1, NCOLL2, …\}. \\ Generate the final sentence only \\ without any comments nor notes. \\ \{EXAMPLES\} \\ \{ORIGINAL SENTENCE\} →} \\ \hline
\makecell{CORR}                        & \makecell{Correct grammar in this French sentence. \\ Generate the final sentence only \\ without any comments nor notes. \\ \{EXAMPLES\} \\ \{ORIGINAL SENTENCE\} →}                                                    \\ \hline                                                            
\end{tabular}%
}
\caption{Content of instructions per type given to Claude 3 Opus}
\label{table:instruc}
\end{table}

\section{Few-shot sentences given to Claude 3 Opus}

Tables \ref{table:instruc-anx-bd} and \ref{table:instruc-anx-c} contain the few-shot sentences used respectively for the "BASE" and "DICT" instructions, and the "CORR" instruction. They were formatted as such in the prompt: \newline [Sentence with masculine generic] → [Gender-neutralized sentence].

\begin{table}[h!]
\centering
\resizebox{\columnwidth}{!}{%
\begin{tabular}{@{}cc@{}}
\toprule
\textbf{Sentence with masculine generic}                                                     & \textbf{Gender-neutralized sentence}                                         \\ \midrule
\makecell{Le président de la FIFA Sepp Blatter \\ rejette les accusations \textbf{des manifestants} \\ en les accusant d’opportunisme. \\ (FIFA President Sepp Blatter \\ dismisses \textbf{the protesters}' \\ accusatations as opportunism.)} &
  \makecell{Le président de la FIFA Sepp Blatter \\ rejette les accusations de \textbf{la manifestation} \\ en l’accusant d'opportunisme. \\ (FIFA President Sepp Blatter \\ dismisses \textbf{the protest}'s \\ accusatations as opportunism.)} \\ \hline
\makecell{\textbf{Les auteurs} et \textbf{les spectateurs} \\ ont été satisfaits des réponses \\ des représentants. \\ (\textbf{Authors} and \textbf{spectators} \\ were pleased with \textbf{the} \\ \textbf{representatives}' responses.)} & \makecell{\textbf{L'autorat} et \textbf{le public} \\ ont été satisfaits des réponses \\ de \textbf{la représentation}. \\ (\textbf{The authorship} and \textbf{the audience} \\ were pleased with the \\ \textbf{representation}'s responses.)} \\ \hline
\makecell{Le vicaire général proposa de disperser \\ \textbf{les religieux} dans d'autres maisons de l'ordre \\ et de procéder à la réfection des bâtiments. \\ (The vicar general suggested to disperse \\ \textbf{religious people} to other houses of the order \\ to repair the buildings.)} &
  \makecell{Le vicaire général proposa de disperser \\ \textbf{le couvent} dans d'autres maisons de l'ordre \\ et de procéder à la réfection des bâtiments. \\ (The vicar general suggested to disperse \\ \textbf{the convent} to other houses of the order \\ to repair the buildings.)} \\ \bottomrule
\end{tabular}%
}
\caption{Few-shot sentences for "BASE" and "DICT" instructions. Bold indicates the differences between sentences with MG and gender-neutralized sentences.}
\label{table:instruc-anx-bd}
\end{table}

\begin{table}[h!]
\centering
\resizebox{\columnwidth}{!}{%
\begin{tabular}{@{}cc@{}}
\toprule
\textbf{RBS-generated sentence with errors}                                                     & \textbf{Manual sentence}                                         \\ \midrule
\makecell{Le président de la FIFA Sepp Blatter \\ rejette les accusations de la manifestation \\ en \textbf{les} accusant d’opportunisme. } &
  \makecell{Le président de la FIFA Sepp Blatter \\ rejette les accusations de la manifestation \\ en \textbf{l’}accusant d'opportunisme.} \\ \hline
\makecell{L'autorat et le public \\ \textbf{a} été satisfaits des réponses \\ des la représentation.} & \makecell{L'autorat et le public \\ \textbf{ont} été satisfaits des réponses \\ de la représentation.} \\ \hline
\makecell{Le vicaire \textbf{générale} proposa de disperser \\ le couvent dans d'autres maisons de l'ordre \\ et de procéder à la réfection des bâtiments.} &
  \makecell{Le vicaire \textbf{général} proposa de disperser \\ le couvent dans d'autres maisons de l'ordre \\ et de procéder à la réfection des bâtiments.} \\ \bottomrule
\end{tabular}%
}
\caption{Few-shot sentences for "CORR" instruction. Bold indicates the differences between the RBS-generated ssentences with error and the manual, correct sentences.}
\label{table:instruc-anx-c}
\end{table}

\section{Manual Error Type Labelling Details}
\label{apx:errtypelabel}

\begin{table*}[h!]
\centering
\resizebox{1.1\textwidth}{!}{%
\fontsize{10pt}{12pt}\selectfont
\begin{tabular}{p{3.5cm}p{20cm}}
\toprule
\textbf{Error Type} & \textbf{Description}                                                                                                                                 \\
\midrule
ADJ                 & Errors related to adjective agreement with the modified noun. Past participles used as adjectives are included in this category.                     \\
\addlinespace
CASE (N)            & Errors related to an incorrect use of lowercase/uppercase characters.                                                                                \\
\addlinespace
DET                 & Errors related to determiner agreement with the modified noun.                                                                                       \\
\addlinespace
DET\_COREF          & Errors related to coreferent possessive determiner agreement with the modified noun.                                                                 \\
\addlinespace
ELISION             & Errors related to elision.                                                                                                                           \\
\addlinespace
GEN\_FAILURE (N)    & Errors related to incorrect text-to-text model generation, most particularly with proper nouns or words that are not part of the model's vocabulary. \\
\addlinespace
MISID\_NOUN &   \begin{tabular}[c]{@{}l@{}}Errors occurring when a member noun's form in the collective-member noun dictionary was wrongly detected as a noun in the original sentence,\\ and was thus incorrectly changed into a CN.\end{tabular} \\
\addlinespace
PREP         & Errors related to preposition usage.                                                                               \\
\addlinespace
PRON\_COREF         & Errors related to coreferent pronoun agreement with the modified noun.                                                                               \\
\addlinespace
PUNCT               & Errors related to punctuation (e.g. missing or double spaces).                                                                                                 \\
\addlinespace
SEM                 & Errors occurring when changing the member noun into its CN counterpart leads to an asemantic sentence.                              \\
\addlinespace
SPECIAL\_CHAR (N)   & Errors related to special characters (e.g. accents).                                                                                                 \\
\addlinespace
UNREPLACED          & Errors occurring when the member noun was not replaced with its CN counterpart.                                                         \\
\addlinespace
VERB                & Errors related to verb or auxiliary agreement.                                                                                                       \\
\bottomrule
\end{tabular}%
}
\caption{Error types and descriptions for the RBS and fine-tuned models}
\label{tab:err}
\end{table*}

\begin{table*}[h!]
\centering
\begin{tabular}{l|ccc|ccc}
\toprule
\textbf{Label} & \multicolumn{3}{c|}{\textbf{Wikipedia}} & \multicolumn{3}{c}{\textbf{Europarl}} \\
 & \textbf{Agree} & \textbf{Disagree} & \textbf{Rate (\%)} & \textbf{Agree} & \textbf{Disagree} & \textbf{Rate (\%)} \\
\midrule
ADJ             & 88  & 26  & 77.19 & 56  & 58  & 49.12 \\
PREP            & 24  & 10  & 70.59 & 8   & 6   & 57.14 \\
PRON\_COREF     & 8   & 8   & 50.00 & 46  & 55  & 45.54 \\
SEM             & 128 & 104 & 55.17 & 40  & 109 & 26.85 \\
VERB            & 91  & 45  & 66.91 & 71  & 74  & 48.97 \\
\bottomrule
\end{tabular}
\caption{Main labels interannotator agreement for Wikipedia and Europarl corpora across all 3 non-instruct-based models (RBS, T5, M2M100)}
\label{tab:intant}
\end{table*}

We give additional information about some of the error types below.

The ELISION error is related to how elision works in French: in the sentences that we are modifying, the masculine determiner "le" and the feminine determiner "la" (\emph{the}) should be elided and written as "l'" when the word that follows begins with a vowel or a mute "h".

The MISID\_NOUN error may occur when the form of a member noun shares several different grammatical categories. For example, "jeunes" (\emph{young}), the member noun's form of the CN "jeunesse" (\emph{youth}), can be both a noun and an adjective. When the adjective form was wrongly detected as a noun, it was included in our dataset and produced an ungrammatical result sentence.

Finally, when it comes to the SEM error type, as discussed by \citet{lecolleNomsCollectifsHumains2019}, CNs in French, and more specifically human CNs, feature specific semantic characteristics due to how they are used to group human beings under a common denomination, based for example on their profession ("le professorat" [\emph{professorate}]), their social status ("l'aristocratie" [\emph{the aristocracy}]), or their political leaning ("la gauche" [\emph{the left}]). Combining human CNs with specific verbs or contexts may thus not be considered semantically correct, and may occur when transforming a sentence. We labeled such transformed sentences with this error.

We report the interannotator agreement for main labels for Wikipedia and Europarl corpora annotations in Table \ref{tab:intant}. It should be noted that agreement for the SEM label is the lowest: this is due to the fact that this type of error has a subjectivity componet, insofar as some sentences, depending on the context, may sound semantic to one person but asemantic to another one. This leads to a difficulty of pinpointing accurately what parts of the sentence may sound odd to a native speaker or may not convey the intended meaning. Moreover, the fact that CNs are relatively uncommon in everyday language and are not as popular as other techniques for gender neutralization makes undiscussed agreement even more difficult. For other labels, interannotator agreement may also be affected by the subtle nature of the changes between sentence pairs (in many cases, the differences are minimal and not immediately perceptible) as well as classification disagreements (past participles categorized as VERB vs. ADJ for example) that were later resolved. All applied labels for all sentences and models were agreed upon after discussion.

\clearpage

\onecolumn

\subsection{Labelling Instructions Given to Annotators (French)}
\label{apx:manualerr}

\begin{spverbatim}
La tâche consiste à annoter les erreurs présentes dans les trois types de phrases générées (par le système à base de règles, T5 et M2M100) pour les deux corpus (Wikipédia et Europarl). Une même phrase peut être annotée avec plusieurs types d'erreurs différents. Si une erreur est répétée, un seul label de même type est appliqué. Les types d'annotations d'erreurs sont définis ci-dessous :

ADJ : erreurs qui portent sur les adjectifs (accord) (p. ex. *« les soldats allemand »)

CASE : erreurs qui portent sur l'utilisation des majuscules/minuscules (p. ex., nom propre non mis en majuscule)

DET : erreurs qui portent sur les déterminants (p. ex. *« le police »)

DET_COREF : erreurs survenues lorsqu’un déterminant possessif qui se rapporte au nom collectif n’a pas été modifié ou a été modifié à tort, (p. ex. « Il a rejoint la police dans leurs locaux » [au lieu de « ses »])

ELISION : erreurs qui portent sur l'élision (p. ex. *« la armée »)

GEN_FAILURE : erreurs qui émanent d'une mauvaise génération du modèle, notamment par exemple sur les noms propres ou les mots inconnus du modèle (p. ex. « Society of Friends » → « Society ofamina »). Dans certains cas, cela peut aussi mener à des déformations de mots par l’ajout ou la suppression d’espaces (p. ex. *« L’organisationPromet »). Dans ce cas, annoter avec GEN_FAILURE et non pas avec SPACE.

MISID_NOUN : cas où une forme présente dans le dictionnaire a été incorrectement identifiée comme nom et donc modifié comme nom collectif, ce qui rend la phrase asémantique/agrammaticale. Dans ce cas, annoter avec MISID_NOUN et non pas avec SEM.

PREP  : erreurs relatives à l'utilisation des propositions (p. ex. *« invasion par de l'armée ennemie »)

PRON_COREF : erreurs survenues lorsqu’un pronom possessif qui se rapporte au nom collectif n’a pas été modifié ou a été modifié à tort (p. ex. « Au gouvernement, les manifestants leur ont exprimé leur colère » [au lieu de « lui »])

PUNCT : erreurs qui portent sur la ponctuation, les espaces

SEM : phrase jugée asémantique après le remplacement du nom de membre par un nom collectif (par exemple, dans des structures numérales de type « l'un des auteurs » → *« l'un de l'autorat » / « les millions de soldats » → ?« les millions de l'armée »)

SPECIAL_CHAR : erreurs qui portent sur les caractères spéciaux (p. ex. accents)

UNREPLACED : cas où le nom de membre n'a pas été remplacé

VERB : erreurs qui portent sur les verbes (conjugaison) ou les auxiliaires (p. ex. *« les soldats parte »)
\end{spverbatim}

\noindent\begin{minipage}{\textwidth}
\section{GPT 4o-mini Automatic Error Type Labelling}
\label{apx:autoerr}

\subsection{Error Type Explanations Generation Prompts}

\begin{lstlisting}[language=xml, caption={System Prompt}, label=lst:system-prompt]
You are an assistant whose task is to quickly summarize error types comparing a golden and an LLM-rewritten sentence.
\end{lstlisting}

\begin{lstlisting}[language=xml, caption={User Prompt}, label=lst:user-prompt]
Given a rewritten golden sentence and an LLM-rewritten sentence, compare the two and write a few words about the errors found in the LLM-rewritten sentence (if any) compared to the golden sentence. Each error type explanation should be separated with a semicolon and written in lowercase. Each explanation should start with a past tense verb. Only output the explanations and keep it short.

<examples>
Golden sentence: [GOLDEN SENTENCE EXAMPLE]
LLM-rewritten sentence: [LLM-REWRITTEN SENTENCE EXAMPLE]
Error type explanations: [EXPLANATION EXAMPLE]

Golden sentence: [GOLDEN SENTENCE EXAMPLE]
LLM-rewritten sentence: [LLM-REWRITTEN SENTENCE EXAMPLE]
Error type explanations: [EXPLANATION EXAMPLE]

Golden sentence: [GOLDEN SENTENCE EXAMPLE]
LLM-rewritten sentence: [LLM-REWRITTEN SENTENCE EXAMPLE]
Error type explanations: [EXPLANATION EXAMPLE]
</examples>

<info>
For your information, the LLM-rewritten sentence was generated after the LLM received this prompt: [PROMPT GIVEN TO INSTRUCTION-TUNED MODEL]
</info>

Golden sentence: [GOLDEN SENTENCE EXAMPLE]
LLM-rewritten sentence: [LLM-REWRITTEN SENTENCE EXAMPLE]
Error type explanations:
\end{lstlisting}
\end{minipage}

\clearpage

\noindent\begin{minipage}{\textwidth}
\subsection{Error Labels Generation Prompts}

\begin{lstlisting}[language=xml, caption={System Prompt}, label=lst:system-prompt]
You are an assistant whose task is to generate short error labels based on error descriptions.
\end{lstlisting}

\begin{lstlisting}[language=xml, caption={User Prompt}, label=lst:user-prompt]
Given a rewritten golden sentence, an LLM-rewritten sentence and an error type explanation, generate error labels for the LLM-rewritten sentence. Each error label should be separated with a semicolon (without spaces) and written in uppercase. If no error is found, output nothing. Only output the labels.

For reference, the previous labels have already been generated: [PREVIOUSLY GENERATED LABELS, IF ANY]
If any of the labels corresponds to the error type explanation, use it. Otherwise, feel free to generate a new label.

<examples>
Golden sentence: [GOLDEN SENTENCE EXAMPLE]
LLM-rewritten sentence: [LLM-REWRITTEN SENTENCE EXAMPLE]
Error type explanations: [EXPLANATION EXAMPLE]
Error labels: [LABEL EXAMPLES]

Golden sentence: [GOLDEN SENTENCE EXAMPLE]
LLM-rewritten sentence: [LLM-REWRITTEN SENTENCE EXAMPLE]
Error type explanations: [EXPLANATION EXAMPLE]
Error labels: [LABEL EXAMPLES]

Golden sentence: [GOLDEN SENTENCE EXAMPLE]
LLM-rewritten sentence: [LLM-REWRITTEN SENTENCE EXAMPLE]
Error type explanations: [EXPLANATION EXAMPLE]
Error labels: [LABEL EXAMPLES]
</examples>

<info>
For your information, the LLM-rewritten sentence was generated after the LLM received this prompt: [PROMPT GIVEN TO INSTRUCT-BASED MODEL]
</info>

Golden sentence: [GOLDEN SENTENCE EXAMPLE]
LLM-rewritten sentence: [LLM-REWRITTEN SENTENCE EXAMPLE]
Error type explanations: [EXPLANATION EXAMPLE]
Error labels:
\end{lstlisting}
\end{minipage}

\clearpage

\twocolumn
\section{Generation Examples}
\label{apx:examples}

\begin{exe}
\ex \begin{xlist}
        \ex Cette démarche fera progresser les droits \textbf{des citoyens}, car, par l'intermédiaire du Parlement, \textbf{les citoyens seront} en contact direct avec la Commission, ce qui lui confèrera une légitimité considérable. [original sent.] \newline
        \begin{small}(This approach will increase \textbf{citizens' [masc.]} rights, because, through the Parliament, \textbf{citizens will [pl.]} have a direct line to the Commission thereby generating considerable legitimacy.)\end{small}
        \ex Cette démarche fera progresser les droits \textbf{de la citoyenneté}, car, par l'intermédiaire du Parlement, \textbf{la citoyenneté seront} en contact direct avec la Commission, ce qui lui confèrera une légitimité considérable. [GeNRe-RBS] \newline
        \begin{small}(This approach will increase the rights of \textbf{the citizenry}, because, through the Parliament, \textbf{the citizenry will [pl.]} have a direct line to the Commission thereby generating considerable legitimacy.)\end{small}
        \ex Cette démarche fera progresser les droits \textbf{de la citoyenneté}, car, par l'intermédiaire du Parlement, \textbf{la citoyenneté sera} en contact direct avec la Commission, ce qui lui confèrera une légitimité considérable. [manual sent.] \newline
        \begin{small}(This approach will increase the rights of \textbf{the citizenry}, because, through the Parliament, \textbf{the citizenry will [sg.]} have a direct line to the Commission thereby generating considerable legitimacy.)\end{small}
    \end{xlist}
    \label{ex1}
\end{exe}

\begin{exe}
\ex \begin{xlist}
        \ex Je vous invite à informer \textbf{les députés européens chargés} des dossiers agricoles de l'avancement des négociations. [original sent.] \newline
        \begin{small}(I urge you to inform \textbf{the Members of European Parliament [masc] in charge of [pl.]} the agricultural issues about the progress of negotiations.)\end{small}
        \ex Je vous invite à informer \textbf{le parlement européen chargés} des dossiers agricoles de l'avancement des négociations. [GeNRe-RBS] \newline
        \begin{small}(I urge you to inform \textbf{the European parliament in charge of [pl.]} the agricultural issues about the progress of negotiations.)\end{small}
        \ex Je vous invite à informer \textbf{le parlement européen chargé} des dossiers agricoles de l'avancement des négociations. [manual sent.] \newline
        \begin{small}(I urge you to inform \textbf{the European parliament in charge of [sg.]} the agricultural issues about the progress of negotiations.)\end{small}
    \end{xlist}
    \label{ex2}
\end{exe}

\begin{exe}
\ex \begin{xlist}
        \ex Un deuxième élément concerne le soutien apporté à la Commission à \textbf{l'actorat local} qui \textbf{veulent} participer à ces programmes afin d'avoir accès aux sources de financement correspondantes. [GeNRe-RBS] \newline
        \begin{small}(A second factor is the Commission's support for \textbf{local actors [coll. sg.]} who \textbf{want [pl.]} to take part in these programmes, so that they can access the corresponding funding mechanisms.)\end{small}
        \ex Un deuxième élément concerne le soutien apporté à la Commission à \textbf{l'actorat local} qui \textbf{veut} participer à ces programmes afin d'avoir accès aux sources de financement correspondantes. [GeNRe-FT-M2M100] \newline
        \begin{small}(A second factor is the Commission's support for \textbf{local actors [coll. sg.]} who \textbf{want [sg.]} to take part in these programmes, so that they can access the corresponding funding mechanisms.)\end{small}
    \end{xlist}
    \label{ex3}
\end{exe}

\begin{exe}
\ex \begin{xlist}
        \ex Dans une lettre à la \textbf{famille} datée du 13 juin 1861, Zeng Guofan a ordonné à \textbf{ses propres navires} de surveiller les navires commerciaux britanniques après avoir remarqué que des marchands étrangers déchargeaient du riz \textbf{à la rébellion} à Anqing. [GeNRe-RBS] \newline
        \begin{small}(In a letter addressed to the \textbf{family} and dated June 13, 1861, Zeng Guofan ordered \textbf{his own vessels} to monitor British commercial vessels after noticing that foreign sellers were giving rice to \textbf{the rebellion} in Anqing.)\end{small}
        \ex Dans une lettre à la \textbf{parenté} datée du 13 juin 1861, Zeng Guofan a ordonné à \textbf{sa propre flotte} de surveiller les navires commerciaux britanniques après avoir remarqué que des marchands étrangers déchargeaient du riz \textbf{aux rebelles} à Anqing. [Claude 3 Opus-BASE] \newline
        \begin{small}(In a letter addressed to the \textbf{kinfolk} and dated June 13, 1861, Zeng Guofan ordered \textbf{his own fleet} to monitor British commercial vessels after noticing that foreign sellers were giving rice to \textbf{rebels} in Anqing.)\end{small}
    \end{xlist}
    \label{ex5}
\end{exe}

\begin{exe}
\ex \begin{xlist}
        \ex Mais l'armée protestante, toujours agressive, \textbf{restaient} à la charge des habitants et \textbf{constituaient} une lourde charge. [GeNRe-RBS] \newline
        \begin{small}(But the Protestant army, still aggressive, \textbf{remained [pl.]} in the care of the local people and \textbf{constituted [pl.]} a heavy burden.)\end{small}
        \ex Mais l'armée protestante, toujours agressive, \textbf{restait} à la charge des habitants et \textbf{constituait} une lourde charge. [Claude 3 Opus-DICT] \newline
        \begin{small}(But the Protestant army, still aggressive, \textbf{remained [sg.]} in the care of the local people and \textbf{constituted [sg.]} a heavy burden.)\end{small}
    \end{xlist}
    \label{ex7}
\end{exe}

\begin{exe}
\ex \begin{xlist}
        \ex Paradoxalement, cette progression en voix s'accompagne d'un recul en nombre d'élus, du fait de la poussée des candidats indépendants (pour la plupart de la \textbf{représentation} de la communauté kurde) et du CHP. [GeNRe-RBS] \newline
        \begin{small}(Paradoxically, this increase in votes paralleled a decrease in the number of elected representatives due to better results for the independent candidates (most of them \textbf{coming from the representation} of the Kurdish community) and CHP.\end{small}
        \ex Paradoxalement, cette progression en voix s'accompagne d'un recul en nombre d'élus, du fait de la poussée des candidats indépendants (pour la plupart des \textbf{représentants} de la communauté kurde) et du CHP. [Claude 3 Opus-DICT] \newline
        \begin{small}(Paradoxically, this increase in votes paralleled a decrease in the number of elected representatives due to better results for the independent candidates (most of them \textbf{being representatives} of the Kurdish community) and CHP.\end{small}
    \end{xlist}
    \label{ex6}
\end{exe}

\end{document}